\documentclass{article}

\usepackage{microtype}
\usepackage{graphicx}
\usepackage{subcaption}
\usepackage{booktabs} 

\usepackage{hyperref}

\usepackage{float}
\usepackage{tabularx}
\usepackage{array} 

\usepackage[preprint]{icml2026}

\usepackage{amsmath}
\usepackage{amssymb}
\usepackage{mathtools}
\usepackage{amsthm}

\usepackage[capitalize,noabbrev]{cleveref}

\theoremstyle{plain}
\newtheorem{theorem}{Theorem}[section]
\newtheorem{proposition}[theorem]{Proposition}

\theoremstyle{definition}
\newtheorem{definition}[theorem]{Definition}

\theoremstyle{remark}

\usepackage[disable,textsize=tiny]{todonotes}

\icmltitlerunning{How Vision Becomes Language: A Layer-wise Information-Theoretic Analysis of Multimodal Reasoning}

\begin{document}

\twocolumn[
  \icmltitle{How Vision Becomes Language: A Layer-wise Information-Theoretic Analysis of Multimodal Reasoning}

\icmlsetsymbol{equal}{*}

\begin{icmlauthorlist}
  \icmlauthor{Hongxuan Wu}{equal,Duke}
  \icmlauthor{Yukun Zhang}{equal,cuhk}
  \icmlauthor{Xueqing Zhou}{equal,fudan}
\end{icmlauthorlist}

\icmlaffiliation{Duke}{Duke kunshan University, Duke University}
\icmlaffiliation{cuhk}{The Chinese University of Hong Kong, Hong Kong, China}
\icmlaffiliation{fudan}{Fudan University, Shanghai, China}

\icmlcorrespondingauthor{Hongxuan Wu}{hongxuan.wu@dukekunshan.edu.cn}
\icmlcorrespondingauthor{Yukun Zhang}{215010026@link.cuhk.edu.cn}
\icmlcorrespondingauthor{Xueqing Zhou}{19210240101@fudan.edu.cn}
  %


]



\printAffiliationsAndNotice{}  

\begin{abstract}
When a multimodal Transformer answers a visual question, is the prediction driven by visual evidence, linguistic reasoning, or genuinely fused cross-modal computation---and how does this structure evolve across layers? We address this question with a layer-wise framework based on Partial Information Decomposition (PID) that decomposes the predictive information at each Transformer layer into redundant, vision-unique, language-unique, and synergistic components. To make PID tractable for high-dimensional neural representations, we introduce \emph{PID Flow}, a pipeline combining dimensionality reduction, normalizing-flow Gaussianization, and closed-form Gaussian PID estimation. Applying this framework to LLaVA-1.5-7B and LLaVA-1.6-7B across six GQA reasoning tasks, we uncover a consistent \emph{modal transduction} pattern: visual-unique information peaks early and decays with depth, language-unique information surges in late layers to account for roughly 82\% of the final prediction, and cross-modal synergy remains below 2\%. This trajectory is highly stable across model variants (layer-wise correlations $>$0.96) yet strongly task-dependent, with semantic redundancy governing the detailed information fingerprint. To establish causality, we perform targeted Image$\rightarrow$Question attention knockouts and show that disrupting the primary transduction pathway induces predictable increases in trapped visual-unique information, compensatory synergy, and total information cost---effects that are strongest in vision-dependent tasks and weakest in high-redundancy tasks. Together, these results provide an information-theoretic, causal account of how vision becomes language in multimodal Transformers, and offer quantitative guidance for identifying architectural bottlenecks where modality-specific information is lost.
\end{abstract}

\section{Introduction}
\label{sec:introduction}

When a multimodal large language model (MLLM) answers a visual question, what drives the prediction---visual evidence, linguistic reasoning, or a genuinely fused cross-modal computation?  The answer has direct design consequences: if visual evidence is absorbed into language representations early in the network, then improving the visual encoder alone will yield diminishing returns, and the integration mechanism itself becomes the bottleneck.  Despite rapid progress in MLLM capabilities, this question remains open.

Existing interpretability tools offer partial views.  Attention-based analyses reveal \emph{where} information is routed but not \emph{what type} of predictive content is carried.  Probing classifiers diagnose what is linearly decodable at each layer but depend on auxiliary task design and do not decompose how multiple sources combine. Gradient-based attributions measure input sensitivity rather than the internal information structure of representations.  What is missing is a \emph{representation-level} account that, at each layer, quantifies how much predictive information is shared by both modalities, unique to one, or available only through their combination.

We provide such an account using \emph{partial information decomposition} (PID).  Given visual and linguistic representations $X_V$ and $X_L$ at a given layer and a target variable $Y$, PID decomposes the joint predictive information into four non-negative terms:
\begin{align}
I(X_V, X_L;\, Y)
\;=\;
\underbrace{R\vphantom{U_V}}_{\text{redundant}}
\;+\;
\underbrace{U_V}_{\text{vision-unique}}
\;+\;
\underbrace{U_L}_{\text{language-unique}}
\;+\;
\underbrace{S\vphantom{U_V}}_{\text{synergistic}}\,.
\label{eq:pid_decomp_intro}
\end{align}
Tracking the quadruple $(R, U_V, U_L, S)$ across all layers of a Transformer yields an \emph{information trajectory}---a depth-resolved fingerprint of how multimodal evidence is injected, transformed, and consolidated through the network.

Applying this framework to LLaVA-1.5-7B and LLaVA-1.6-7B across six GQA reasoning tasks, we uncover a consistent \emph{modal transduction} pattern: visual-unique information peaks in early layers and decays monotonically with depth, while language-unique information surges in late layers and dominates the final prediction.  Synergy---the information available only when both modalities are considered jointly---remains surprisingly small throughout.  At the decision layer, language-unique information accounts for roughly 82\% of total predictive information, vision-unique contributes approximately 6\%, and synergy stays below 2\%.  This suggests that current MLLMs do not perform ``emergent cross-modal fusion'' so much as \emph{translate} visual evidence into language-space representations for downstream reasoning.

\paragraph{Contributions.}
This paper makes four contributions.

\begin{enumerate}

\item \textbf{A mechanism taxonomy grounded in information trajectories.}
We define three mechanistic regimes---\emph{persistent synergy}, \emph{modal transduction}, and \emph{redundancy-dominant convergence}---as formally separable trajectory signatures (Definitions~\ref{def:persistent_synergy}--\ref{def:redundant_convergence}).  These provide a reusable vocabulary for characterizing cross-modal computation in any architecture that maintains distinguishable modality streams.

\item \textbf{PID Flow: tractable PID estimation for high-dimensional neural representations.}
Direct PID computation on Transformer activations (typically $d \geq 4096$) is infeasible.  We introduce \emph{PID Flow}, a three-stage pipeline---dimensionality reduction, normalizing-flow Gaussianization, and closed-form Gaussian PID---that exploits the bijective invariance of mutual information to yield stable estimates.  The estimator is layer-independent, architecture-agnostic, and requires no task-specific probing design.

\item \textbf{Empirical identification of modal transduction as the dominant regime in LLaVA.}
Across six semantically diverse tasks, both LLaVA-1.5 and LLaVA-1.6 exhibit strikingly consistent transduction trajectories (cross-model layer-wise correlations $> 0.96$), with shared turning points around layers~18--20.  Task semantics modulate the magnitude of redundancy and the strength of language dominance but do not alter the overarching regime---modal transduction is architecture-robust and task-general within the LLaVA family.

\item \textbf{Causal validation via attention knockout.}
We block the Image$\rightarrow$Question attention pathway and show that the resulting PID shifts match three \emph{a priori} predictions of the transduction hypothesis: vision-unique information increases (visual evidence becomes ``trapped''), synergy rises (the model resorts to less efficient joint processing), and the total information budget grows (compensation is costly).  Task-dependent effect sizes further reveal that semantic redundancy governs robustness to pathway disruption, providing the first information-theoretic, causal account of attention-knockout effects in multimodal Transformers.

\end{enumerate}

Beyond interpretation, PID trajectories offer actionable architectural guidance.  If visual information is compressed in early layers, encoder improvements alone cannot change the downstream computation; preserving more $U_V$ likely requires objective- or architecture-level interventions such as auxiliary losses that penalise premature transduction or cross-modal attention mechanisms that maintain a separate visual stream.  More broadly, PID trajectories can identify bottleneck layers where modality-unique information is lost, guiding targeted modifications rather than wholesale redesign.

The remainder of the paper is organized as follows.  Section~\ref{sec:related_work} reviews related work.  Section~\ref{sec:prelim_pid} introduces partial information decomposition.  Section~\ref{sec:method} presents the layer-wise PID framework and the PID Flow estimator.  Section~\ref{sec:experiments} reports experiments and causal interventions.  Section~\ref{sec:conclusion} concludes.

\begin{figure*}[t]
    \centering
    \includegraphics[width=\textwidth]{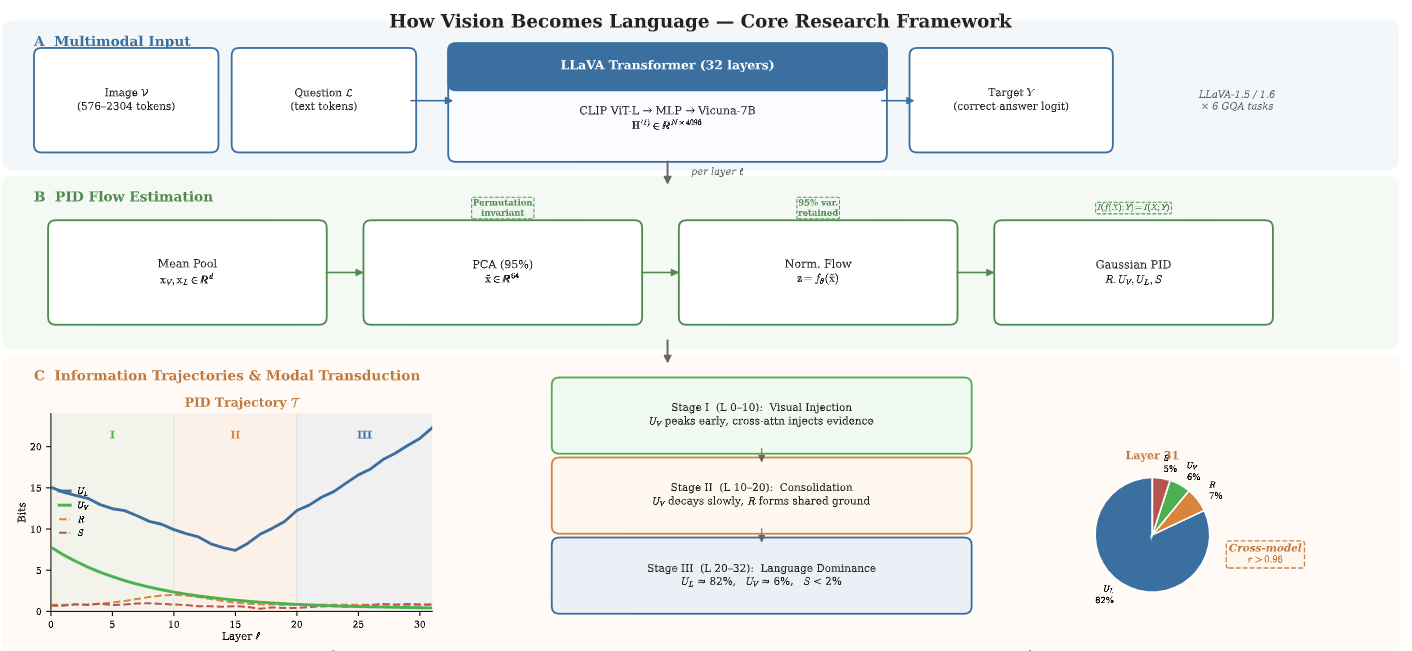}
    \caption{%
    \textbf{Core research framework.}
    \textbf{(A)}~Multimodal input: image and question tokens are processed by a 32-layer LLaVA Transformer.
    \textbf{(B)}~PID Flow estimation: layer-wise representations are compressed via mean pooling and PCA, Gaussianized by a normalizing flow, and decomposed into four PID components ($R, U_V, U_L, S$).
    \textbf{(C)}~Information trajectories reveal a three-stage \emph{modal transduction} pattern: visual injection (Stage~I), consolidation (Stage~II), and language-dominant decision formation (Stage~III), with $U_L \approx 82\%$ and $S < 2\%$ at the final layer.
    Cross-model trajectory correlations exceed $r > 0.96$.}
    \label{fig:framework}
\end{figure*}

\section{Related Work}
\label{sec:related_work}

Modern vision--language models (VLMs) combine a vision encoder with a large
language model (LLM) backbone through trainable adapters.
CLIP~\cite{radford2021learning} established strong cross-modal representations
via contrastive pre-training, and instruction-tuned systems such as
LLaVA~\cite{liu2023visual} further improved performance on downstream
vision--language tasks.
Despite these advances, the \emph{mechanisms} of multimodal reasoning remain
unclear: when an MLLM answers a visual question, which parts of the computation
depend on vision, which depend on language, and where (if anywhere) does
genuinely joint cross-modal evidence emerge?
We review three lines of work most relevant to our goal---mechanistic
interpretability via attention, representation-level probing and
information-theoretic diagnostics, and partial information decomposition (PID),
together with causal interventions through attention knockout.

\paragraph{Attention-based interpretability.}
A large body of work analyzes self- and cross-attention patterns in Transformers
as a proxy for information routing.
Tooling such as VLM-Interpret visualizes attention maps and relevance scores in
LLaVA-style models, enabling interactive inspection of how image patches and
text tokens are attended during reasoning~\citep{benmelech2024lvmlinterpret}.
Beyond raw attention rollout, GMAR proposes a gradient-weighted multi-head
attention rollout that re-weights heads by class-specific gradient importance,
yielding sharper token-level saliency in Vision Transformers~\citep{jo2025gmar}.
Earlier analyses in language-only Transformers also show that some attention
heads correlate with syntactic or semantic relations, while many heads appear
redundant or allocate mass to special tokens~\citep{clark2019does}.
However, attention weights can be weakly correlated with feature importance
measured by gradients or erasure, and different attention patterns may lead to
similar predictions, raising concerns about using attention alone as a faithful
explanation~\citep{jain2019attention}.
In the visual domain, relevance-propagation schemes that combine attention with
gradient information (e.g., Deep Taylor-style decompositions) produce
class-specific attribution maps that better align with human rationales for
images and text~\citep{chefer2021transformer}.
In multimodal settings, fine-grained attention analyses together with causal
interventions on visual features further highlight the gap between attention
heatmaps and true decision mechanisms~\citep{bi2025unveiling}.
Overall, attention-based methods primarily answer \emph{where} the model routes
information, but they do not directly quantify \emph{what type} of predictive
information is carried (redundant, unique, or synergistic), which motivates our
information-structural approach.

\paragraph{Probing and information-theoretic diagnostics.}
Probing methods diagnose representation content by training auxiliary models on
intermediate activations.
Linear probes provide simple layer-wise diagnostics of linear separability and
representation accessibility~\citep{alain2016linear}.
Structural and edge probing demonstrate that contextual encoders implicitly
encode rich linguistic structure, recovering parse trees and diverse syntactic
phenomena from frozen representations~\citep{hewitt2019structural,tenney2019what}.
Recent work formalizes probing in information-theoretic terms, treating probe
accuracy as evidence about how much information a representation contains about
a target and emphasizing controls and baselines to avoid confounds (e.g.,
lexical shortcuts)~\citep{pimentel2020information,pimentel2021bayesian}.
Complementary to supervised probes, sparse autoencoder approaches learn
interpretable feature dictionaries directly from activations; in the multimodal
setting, SAE-based analyses can disentangle monosemantic features and study
feature-level alignment and restructuring in VLMs~\citep{lou2025saev}.
These methods are valuable for \emph{what is encoded}, but they typically do not
decompose \emph{how multiple sources combine} to form predictions, especially in
the presence of redundancy and synergy.

\paragraph{Partial information decomposition.}
PID decomposes mutual information from multiple sources into redundant, unique,
and synergistic components, providing a principled language for multi-source
predictive structure~\citep{williams2010nonnegative,bertschinger2014quantifying,lyu2024explicit,murphy2024information}.
Exact PID is generally intractable for continuous high-dimensional variables,
so recent work has focused on tractable estimators and relaxations, including
Gaussian PID in closed form~\citep{venkatesh2023gaussian}, flow-based mappings to
latent Gaussian spaces~\citep{zhao2025partial}, and dynamic extensions based on
information rates for multivariate processes~\citep{faes2025partial}, as well as
geometric viewpoints clarifying relationships among PID solutions and dependency
structure~\citep{kunert2020partial}.
PID has been used to quantify multimodal interactions~\citep{liang2023quantifying},
interpret diffusion-based text-to-image systems~\citep{zawar2024diffusionpid},
and design interaction-aware measures and objectives for multimodal learning and
mixture-of-experts settings~\citep{yang2025efficient,dissanayake2025quantifying,xin2025I2MoE}.
In contrast to most prior applications that focus on distribution-level summaries
or specialized architectures, our work uses PID \emph{layer-wise} to characterize
the depth-wise evolution of information structure and to identify mechanistic
regimes in large pretrained vision--language Transformers.

\paragraph{Causal interventions via attention knockout.}
To move from correlational patterns to causal mechanism testing, we adopt
attention knockout as an intervention on information routing.
In language-only LLMs, Geva et al.~\cite{geva2023dissecting} ablate specific
attention connections to localize where factual knowledge is read out, measuring
the impact on predicted answers.
We extend this idea to multimodal models by selectively blocking attention from
image tokens to question tokens (Image$\rightarrow$Question), and analyzing how
both prediction behavior and information structure change.
Crucially, we pair knockout with PID trajectories: knockout provides a causal
perturbation, while PID quantifies how the model reallocates predictive
information across redundancy, modality-unique evidence, and synergy under
intervention.

\textbf{From prior work to our approach.}
Taken together, existing lines of research provide complementary but incomplete views of multimodal reasoning. Attention-based analyses reveal \emph{where} information may flow but do not characterize the \emph{type} of predictive information being transferred; probing and representation diagnostics identify \emph{what} is encoded at individual layers but largely ignore how multiple modalities combine; and prior applications of PID quantify multimodal interaction at an aggregate level, without resolving depth-wise dynamics or mechanistic regimes. Moreover, causal interventions such as attention knockout have been used to localize important pathways, but lack a principled account of \emph{why} disrupting a pathway affects performance. In the next section, we introduce a layer-wise PID framework that bridges these gaps by jointly tracking the evolution of redundant, unique, and synergistic information across depth, and by pairing this analysis with targeted attention knockouts to obtain causal, information-theoretic explanations of multimodal computation.

\section{Preliminaries: Partial Information Decomposition}
\label{sec:prelim_pid}

Our analysis builds on \emph{Partial Information Decomposition} (PID), a
framework that characterizes how multiple information sources jointly
contribute to a prediction.
Unlike mutual information, which quantifies only the total dependence between
representations and outputs, PID explicitly distinguishes \emph{redundant},
\emph{unique}, and \emph{synergistic} contributions.
This distinction is essential for understanding multimodal reasoning, where
visual and linguistic representations may overlap in meaning, contribute
independently, or interact in a genuinely joint manner.

\subsection{PID formulation}

\paragraph{Beyond mutual information.}
Mutual information $I(X;Y)$ measures how informative a variable $X$ is about a
target $Y$, but it is insufficient when multiple sources are involved.
Given two sources $X_V$ (vision) and $X_L$ (language), $I(X_V,X_L;Y)$ does not
reveal whether predictive information is shared by both modalities, specific to
one modality, or only available through their combination.
This limitation is particularly salient in vision--language models, where
answers may rely on aligned concepts, modality-specific cues, or cross-modal
binding.

\paragraph{Decomposition.}
PID decomposes the joint mutual information into four non-negative components:
\begin{equation}
I(X_V, X_L; Y) = R + U_V + U_L + S ,
\label{eq:pid_decomposition_main}
\end{equation}
where $R$ denotes \emph{redundant information} shared by both sources, $U_V$ and
$U_L$ denote \emph{vision-unique} and \emph{language-unique} information,
respectively, and $S$ denotes \emph{synergistic information} that is available
only when both sources are considered jointly.
These components satisfy consistency relations with marginal and conditional
mutual information (e.g., $I(X_V;Y)=R+U_V$ and $I(X_V;Y\mid X_L)=U_V+S$), ensuring
that the decomposition is additive and interpretable.

\subsection{Interpretation in multimodal models}

\paragraph{Semantic roles.}
In multimodal architectures, PID components admit a natural interpretation.
Redundancy $R$ captures aligned or overlapping concepts represented in both
vision and language (e.g., common object categories).
Vision-unique information $U_V$ corresponds to visual attributes such as color,
position, or fine-grained appearance that are not recoverable from text alone.
Language-unique information $U_L$ reflects linguistic structure, compositional
logic, or task instructions.
Synergy $S$ captures genuinely joint evidence, such as binding an attribute to a
specific object or resolving relations that require simultaneous access to both
modalities.
Table~\ref{tab:pid_semantics} summarizes these roles.

\begin{table}[t]
\centering
\caption{Semantic interpretation of PID components in multimodal models.}
\label{tab:pid_semantics}
\small
\setlength{\tabcolsep}{6pt}
\begin{tabular}{c p{0.28\linewidth} p{0.42\linewidth}}
\toprule
Component & Information-theoretic meaning & Multimodal interpretation \\
\midrule
$R$   & shared predictive information
     & aligned vision--language concepts \\
$U_V$ & information unique to vision
     & visual attributes (color, position, texture) \\
$U_L$ & information unique to language
     & linguistic or logical structure \\
$S$   & information available only jointly
     & cross-modal binding (e.g., object--attribute relations) \\
\bottomrule
\end{tabular}
\end{table}

\paragraph{Expected depth-wise behavior.}
Applied layer-wise to a multimodal Transformer, PID provides a language for
describing how information structure evolves with depth.
A priori, one may expect early layers to contain substantial vision-unique
information, intermediate layers to increase redundancy as modalities align,
and late layers to become dominated by language-unique information if visual
evidence has been successfully transduced into the language representation.
These expectations motivate our empirical analysis of PID trajectories across
layers.

\subsection{Why estimation is non-trivial}

\paragraph{High-dimensional challenge.}
Directly estimating PID from high-dimensional neural representations is
statistically challenging.
Transformer activations typically have thousands of dimensions, while available
sample sizes are limited, making naive density estimation or discretization
infeasible.
Moreover, many existing estimators target mutual information rather than the
full PID structure.

\paragraph{Approach overview.}
To make layer-wise PID analysis tractable in modern vision--language models, we
adopt a strategy that combines dimensionality reduction, invertible
transformations, and closed-form Gaussian information measures.
The goal is not to recover an exact ground-truth decomposition, but to obtain a
stable and interpretable \emph{information-structural proxy} that can be tracked
across depth and compared under causal interventions.
Full methodological and implementation details are provided in
Section~\ref{sec:method} and the Appendix.

\paragraph{Summary.}
PID provides a principled framework for decomposing predictive information into
redundant, modality-unique, and synergistic components.
In the following sections, we apply PID layer-wise to large vision--language
models and use it to identify consistent mechanistic regimes and to interpret
the effects of causal interventions such as attention knockout.

\section{Method: Layer-wise PID Analysis}
\label{sec:method}

We propose a layer-wise analysis framework that quantifies how visual and
linguistic evidence is organized across the depth of a multimodal large
language model (MLLM).
The method proceeds in four steps.
We first define a \emph{layer-wise information state}
(Section~\ref{sec:layerwise_state}),
then operationalize distinct cross-modal mechanisms via trajectory-level
signatures (Section~\ref{sec:three_mechanisms}),
introduce \emph{PID Flow}, a scalable estimator for high-dimensional
representations (Section~\ref{sec:pid_flow}),
and finally discuss estimation reliability and causal diagnostics
(Section~\ref{sec:reliability}).

\subsection{Layer-wise Information State}
\label{sec:layerwise_state}

\paragraph{Setup.}
Consider a multimodal Transformer with $L$ blocks.
We index layers such that $\ell=0$ corresponds to the embedding layer (before
any Transformer block), and $\ell=1,\dots,L$ correspond to the outputs after
each block.
For a given input, token positions are partitioned into two index sets:
the visual region $\mathcal{V}$ (image tokens) and the language region
$\mathcal{L}$ (question tokens).
Let $\mathbf{H}^{(\ell)} \in \mathbb{R}^{N \times d}$ denote the hidden-state
matrix after layer $\ell$, where $N$ is the sequence length and $d$ is the
hidden dimension.

\paragraph{Modality-level summaries.}
We summarize each modality by mean pooling over its token region:
\begin{align}
\mathbf{x}_V^{(\ell)}
&= \frac{1}{|\mathcal{V}|}\sum_{i \in \mathcal{V}} \mathbf{h}_i^{(\ell)}, \qquad
\mathbf{x}_L^{(\ell)}
= \frac{1}{|\mathcal{L}|}\sum_{j \in \mathcal{L}} \mathbf{h}_j^{(\ell)},
\label{eq:mean_pooling}
\end{align}
where $\mathbf{h}_i^{(\ell)}$ denotes the $i$-th token representation at layer
$\ell$.
Mean pooling is a simple permutation-invariant summary; alternative pooling
schemes yield qualitatively similar trajectories
(Appendix~\ref{app:pooling}).

\paragraph{Information state.}
Let $Y$ denote the (discrete) target variable used for evaluation.
We define the \emph{information state} at layer $\ell$ as the PID quadruple
\begin{align}
\mathcal{I}^{(\ell)}
\;=\;
\bigl(R^{(\ell)},\, U_V^{(\ell)},\, U_L^{(\ell)},\, S^{(\ell)}\bigr),
\label{eq:info_state}
\end{align}
induced by the joint predictive information
\begin{align}
I\!\left(\mathbf{x}_V^{(\ell)}, \mathbf{x}_L^{(\ell)}; Y\right)
=
R^{(\ell)} + U_V^{(\ell)} + U_L^{(\ell)} + S^{(\ell)} .
\label{eq:pid_decomp_layer}
\end{align}

\paragraph{Trajectory.}
The layer-wise PID trajectory is the sequence
\begin{align}
\mathcal{T}
\;=\;
\left\{\mathcal{I}^{(\ell)}\right\}_{\ell=0}^{L}.
\label{eq:pid_trajectory}
\end{align}
This trajectory serves as a quantitative information trace describing how
multimodal evidence is injected, transformed, and consolidated throughout the
network.

Table~\ref{tab:pid_components_semantics} summarizes the PID components and their
interpretation in multimodal large language models.

\begin{table}[t]
\centering
\caption{PID components at layer $\ell$ and their interpretation in MLLMs.}
\label{tab:pid_components_semantics}
\small
\setlength{\tabcolsep}{6pt}
\begin{tabular}{l c p{0.52\linewidth}}
\toprule
Component & Symbol & Interpretation (MLLMs) \\
\midrule
Redundancy        & $R^{(\ell)}$
& predictive information jointly encoded by both modalities \\
Vision-unique     & $U_V^{(\ell)}$
& evidence available only from the visual representation \\
Language-unique   & $U_L^{(\ell)}$
& evidence available only from the language representation \\
Synergy           & $S^{(\ell)}$
& evidence available only from the joint pair (PID sense) \\
\bottomrule
\end{tabular}
\end{table}

\subsection{Mechanism Taxonomy via Trajectory Signatures}
\label{sec:three_mechanisms}

We characterize cross-modal computation through trajectory-level signatures
defined on measurable PID quantities, without assuming linearity or
input-independent attention.

\paragraph{Total predictive information.}
We define the total predictive information at layer $\ell$ as
\begin{align}
I_{\mathrm{tot}}^{(\ell)}
\;=\;
R^{(\ell)} + U_V^{(\ell)} + U_L^{(\ell)} + S^{(\ell)} .
\label{eq:Itotal_def}
\end{align}

\paragraph{Persistent synergy.}
\begin{definition}[Persistent synergy]
\label{def:persistent_synergy}
An MLLM exhibits \emph{persistent synergy} if there exist thresholds
$\tau_S>0$ and $\gamma\in(0,1)$ such that
\begin{align}
S^{(L)} > \tau_S
\quad\text{and}\quad
\frac{S^{(L)}}{I_{\mathrm{tot}}^{(L)}} > \gamma .
\label{eq:def_persistent_synergy}
\end{align}
\end{definition}
This condition indicates that the final decision relies on predictive
information not recoverable from either modality alone.

\paragraph{Modal transduction.}
\begin{definition}[Modal transduction]
\label{def:modal_transduction}
An MLLM exhibits \emph{modal transduction} if there exists a turning layer
$\ell^\star\in\{1,\dots,L-1\}$ and a threshold $\eta\in(0,1)$ such that
\begin{align}
U_V^{(\ell)} \text{ is unimodal with a peak near } \ell^\star,
\qquad
\frac{U_L^{(L)}}{I_{\mathrm{tot}}^{(L)}} > \eta .
\label{eq:def_modal_transduction}
\end{align}
\end{definition}
Operationally, this corresponds to early injection of vision-unique evidence
followed by late-layer dominance of language-unique evidence.

\paragraph{Redundancy-dominant convergence.}
\begin{definition}[Redundancy-dominant convergence]
\label{def:redundant_convergence}
An MLLM exhibits \emph{redundancy-dominant convergence} if there exist
$\ell_0\in\{0,\dots,L-1\}$ and $\rho>1$ such that
\begin{align}
R^{(L)} > R^{(\ell_0)}
\quad\text{and}\quad
R^{(L)} >
\rho\cdot\max\!\left\{U_V^{(L)},\, U_L^{(L)},\, S^{(L)}\right\}.
\label{eq:def_redundant_competition}
\end{align}
\end{definition}

Thresholds are treated as descriptive hyperparameters rather than decision
boundaries; all qualitative conclusions are robust across a wide range of
values (Appendix~\ref{app:threshold_sensitivity}).

\begin{proposition}[Separability of mechanisms]
\label{prop:mechanism_separability}
Under non-degenerate conditions
($I_{\mathrm{tot}}^{(L)}>0$ and at least one PID component strictly dominates),
the three definitions above describe mutually exclusive regimes in the space
of final-layer allocations.
\end{proposition}
A proof sketch is provided in Appendix~\ref{app:separability}.

\subsection{PID Flow: Scalable Estimation for High-Dimensional States}
\label{sec:pid_flow}

\paragraph{Overview.}
Direct PID estimation on Transformer representations is infeasible due to
high dimensionality.
We therefore introduce \emph{PID Flow}, a three-stage pipeline combining
dimension reduction, invertible Gaussianization, and closed-form computation
under a Gaussian approximation.

\subsubsection{Step 1: Dimensionality reduction}
We apply PCA to $\mathbf{x}_V^{(\ell)}$ and $\mathbf{x}_L^{(\ell)}$ to obtain
$\tilde{\mathbf{x}}_V^{(\ell)},\tilde{\mathbf{x}}_L^{(\ell)}\in\mathbb{R}^{d'}$,
retaining $95\%$ of the variance in all experiments.
PCA is not invertible; we treat this as a controlled approximation and assess
sensitivity to $d'$ in Appendix~\ref{app:pca_sensitivity}.

\subsubsection{Step 2: Invertible Gaussianization}
On the reduced space, we learn a bijection
$f_\theta:\mathbb{R}^{d'}\to\mathbb{R}^{d'}$ using a normalizing flow.
We define
\begin{align}
\mathbf{z}_V^{(\ell)} = f_\theta(\tilde{\mathbf{x}}_V^{(\ell)}), \qquad
\mathbf{z}_L^{(\ell)} = f_\theta(\tilde{\mathbf{x}}_L^{(\ell)}).
\label{eq:flow_transform}
\end{align}

\begin{theorem}[Bijection invariance of mutual information]
\label{thm:bij_invariance_mi}
Let $f$ be a bijection. Then $I(f(X);Y)=I(X;Y)$ and
$I(f(X_1),f(X_2);Y)=I(X_1,X_2;Y)$.
\end{theorem}

\paragraph{Remark.}
The theorem applies to the flow step. Any PID computed from
$(\mathbf{z}_V^{(\ell)},\mathbf{z}_L^{(\ell)})$ therefore reflects the
information structure of the reduced variables, up to the approximation
introduced by PCA.

\subsubsection{Step 3: Gaussian plug-in PID}
We model $(\mathbf{z}_V^{(\ell)},\mathbf{z}_L^{(\ell)})$ as approximately
Gaussian, estimate parameters
$(\hat{\boldsymbol{\mu}},\hat{\boldsymbol{\Sigma}})$ by maximum likelihood, and
compute PID components in closed form using the $I_{\min}$ redundancy rule
(Appendix~\ref{app:gaussian_pid}).

\subsection{Estimation Reliability and Diagnostics}
\label{sec:reliability}

\paragraph{Consistency.}
\begin{proposition}[Consistency (informal)]
\label{prop:consistency}
If the transformed variables are exactly Gaussian and the redundancy rule is
well-defined, Gaussian plug-in estimates of mutual information and the induced
PID components are consistent as the sample size $n\to\infty$.
\end{proposition}

\paragraph{Attention knockout integration.}
Let $\alpha_{i\to j}$ denote the pre-softmax attention logit from token $i$ to
token $j$.
To probe pathway-level causality, we block a source$\to$target attention pathway
by applying
\begin{align}
\alpha_{i\to j} \leftarrow
\begin{cases}
-\infty, & i\in\mathcal{S},\, j\in\mathcal{T},\\
\alpha_{i\to j}, & \text{otherwise},
\end{cases}
\label{eq:knockout_attention}
\end{align}
and recompute the PID trajectory to analyze shifts
$\Delta\mathcal{I}^{(\ell)}$.

\subsection{Summary}
\label{sec:method_summary}

We introduced a layer-wise PID analysis framework based on
(i) an information-state representation,
(ii) trajectory-defined mechanism signatures, and
(iii) PID Flow, a scalable estimator combining PCA, invertible Gaussianization,
and Gaussian plug-in PID.
The next section applies this framework to LLaVA models and evaluates robustness
across architectures and causal interventions.

\section{Experiments}
\label{sec:experiments}

\paragraph{Goal and design.}
Our experiments use Partial Information Decomposition (PID) to turn
``information structure'' from a descriptive artifact into a set of testable
mechanistic claims about multimodal reasoning.
We organize the evaluation around four questions:
(Q1) \emph{Mechanism identification}---does an MLLM operate closer to persistent
synergy, modal transduction, or redundancy-dominant convergence?
(Q2) \emph{Generality}---is the mechanism consistent across tasks and stable
across model variants?
(Q3) \emph{Causal validation}---do PID components shift in predictable ways when
a hypothesized pathway is disrupted?
(Q4) \emph{Task dependence}---how do tasks differ in their demand for
vision-unique information, redundancy, and synergy?

\paragraph{Experimental suites.}
We conduct three suites of experiments.
First, we compute full 32-layer PID trajectories on six representative GQA-style
reasoning tasks for \textsc{LLaVA-1.5-7B}, and identify the dominant mechanism
via trajectory signatures (Section~\ref{subsec:results_modal_transduction}).
Second, we replicate the pipeline on \textsc{LLaVA-1.6-7B} and quantify
cross-model alignment in both endpoint composition and depth-wise dynamics
(Section~\ref{subsec:results_cross_model}).
Third, we perform attention knockout that blocks the
Image$\rightarrow$Question pathway, and analyze knockout-induced trajectory
shifts to causally validate the transduction hypothesis and expose task-level
heterogeneity (Section~\ref{subsec:results_knockout}).

\paragraph{Operationalization and reliability controls.}
At each layer $\ell$, we aggregate image-token and question-token hidden states
into modality-level representations and define the task label as the target
variable $Y$. We estimate PID components
$\mathcal{I}^{(\ell)} = (R^{(\ell)}, U_V^{(\ell)}, U_L^{(\ell)}, S^{(\ell)})$
and assemble trajectories across depth.
We use (i) cross-model trajectory correlation, (ii) stability of key turning
layers, and (iii) knockout-induced relative changes as primary metrics.
All details on data filtering, representation extraction, estimator settings,
hyperparameters, and statistical tests are deferred to the appendix.

\subsection{Results I: Layer-wise Evidence for Modal Transduction}
\label{subsec:results_modal_transduction}

\begin{figure*}[t]
    \centering
    \begin{subfigure}[t]{0.32\linewidth}
        \centering
        \includegraphics[width=\linewidth]{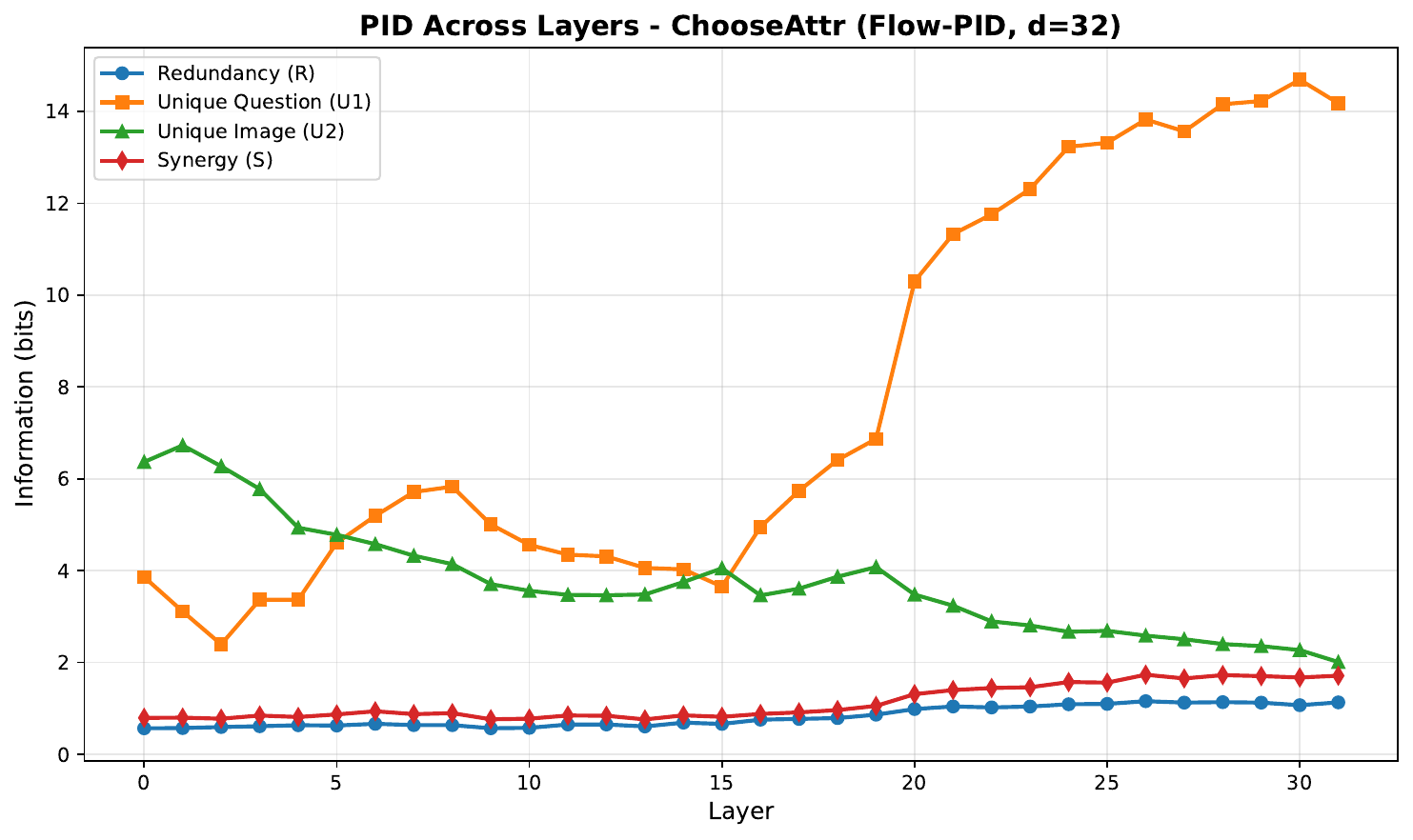}
        \caption{Choose Attribute}
    \end{subfigure}
    \hfill
    \begin{subfigure}[t]{0.32\linewidth}
        \centering
        \includegraphics[width=\linewidth]{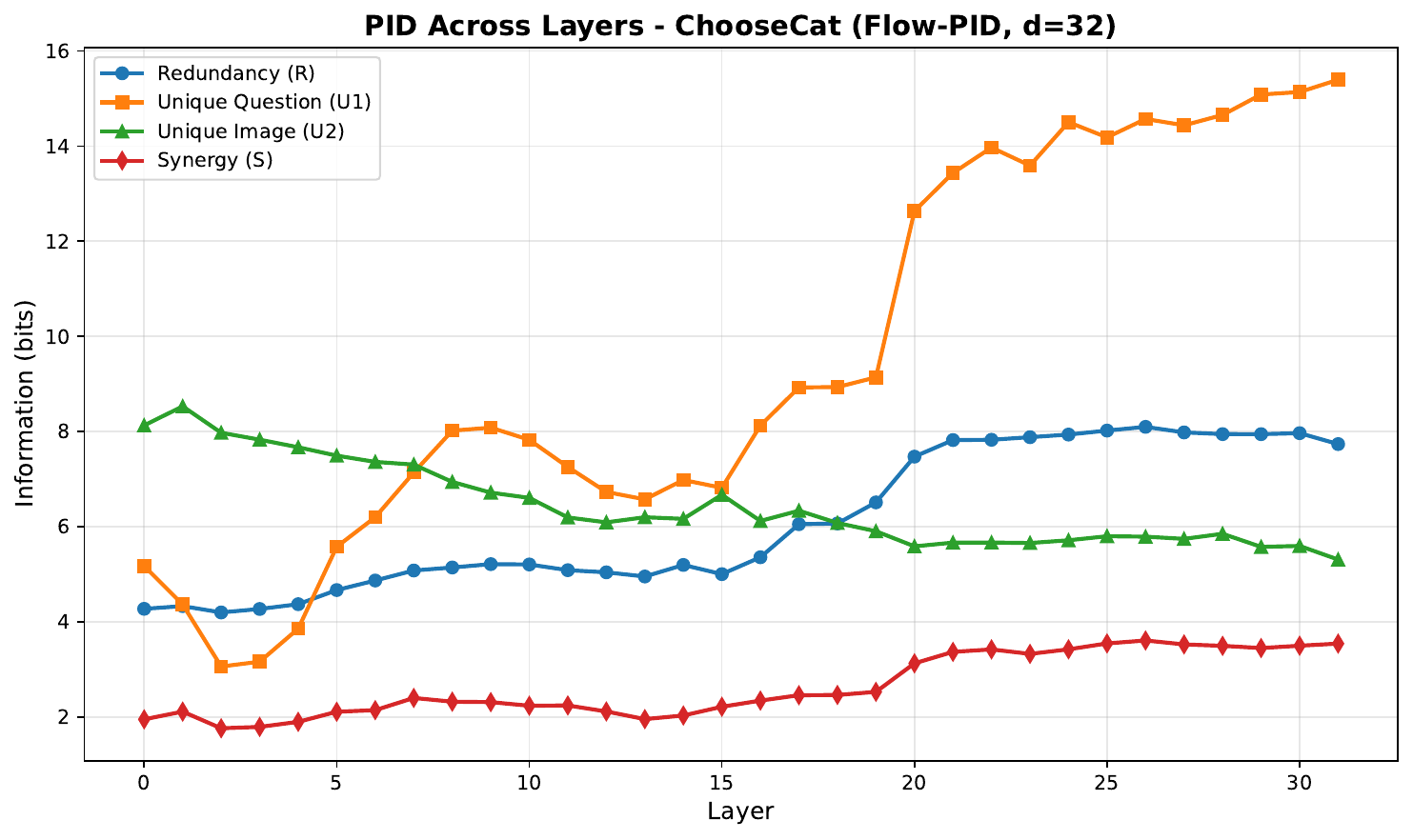}
        \caption{Choose Category}
    \end{subfigure}
    \hfill
    \begin{subfigure}[t]{0.32\linewidth}
        \centering
        \includegraphics[width=\linewidth]{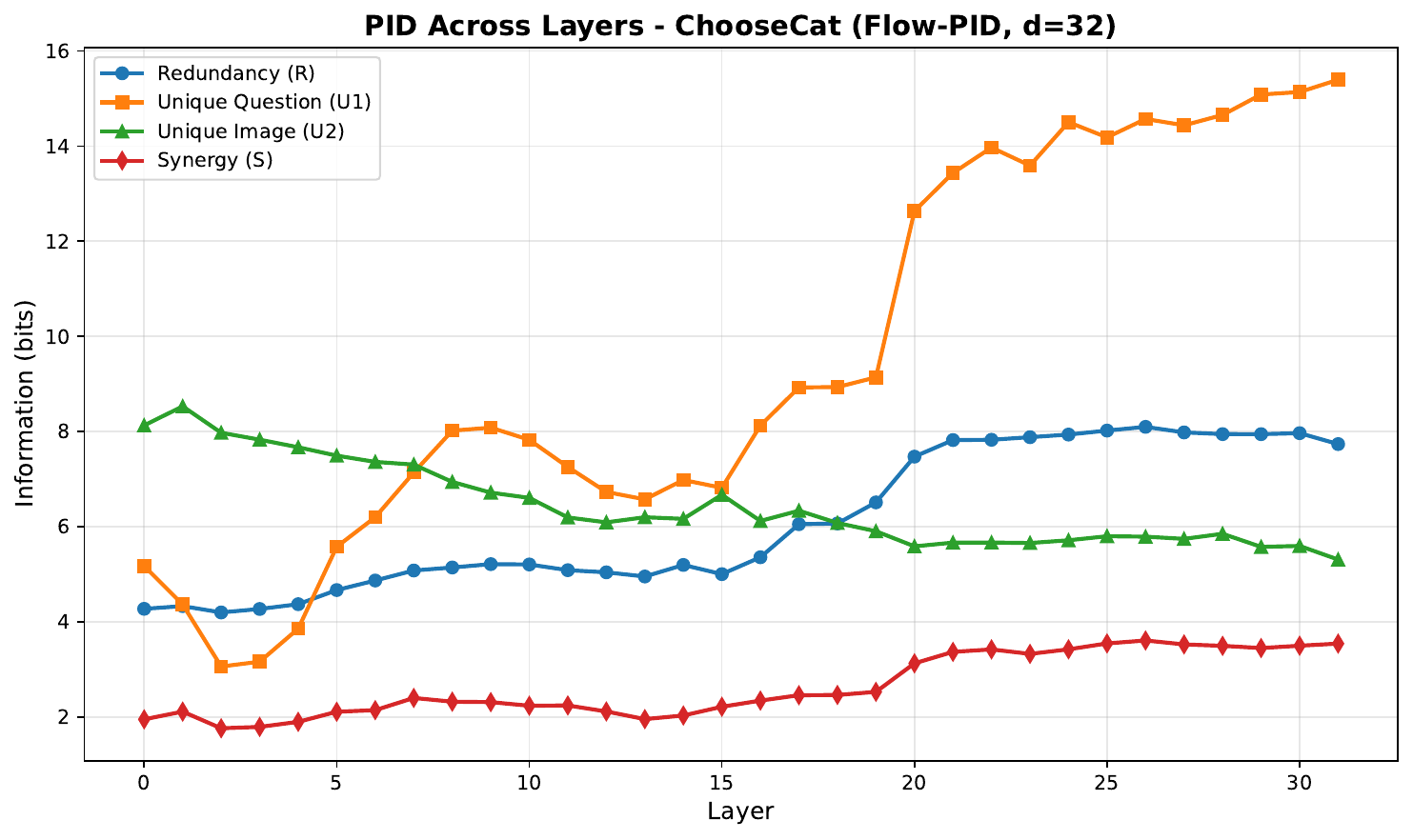}
        \caption{Choose Relation}
    \end{subfigure}

    \begin{subfigure}[t]{0.32\linewidth}
        \centering
        \includegraphics[width=\linewidth]{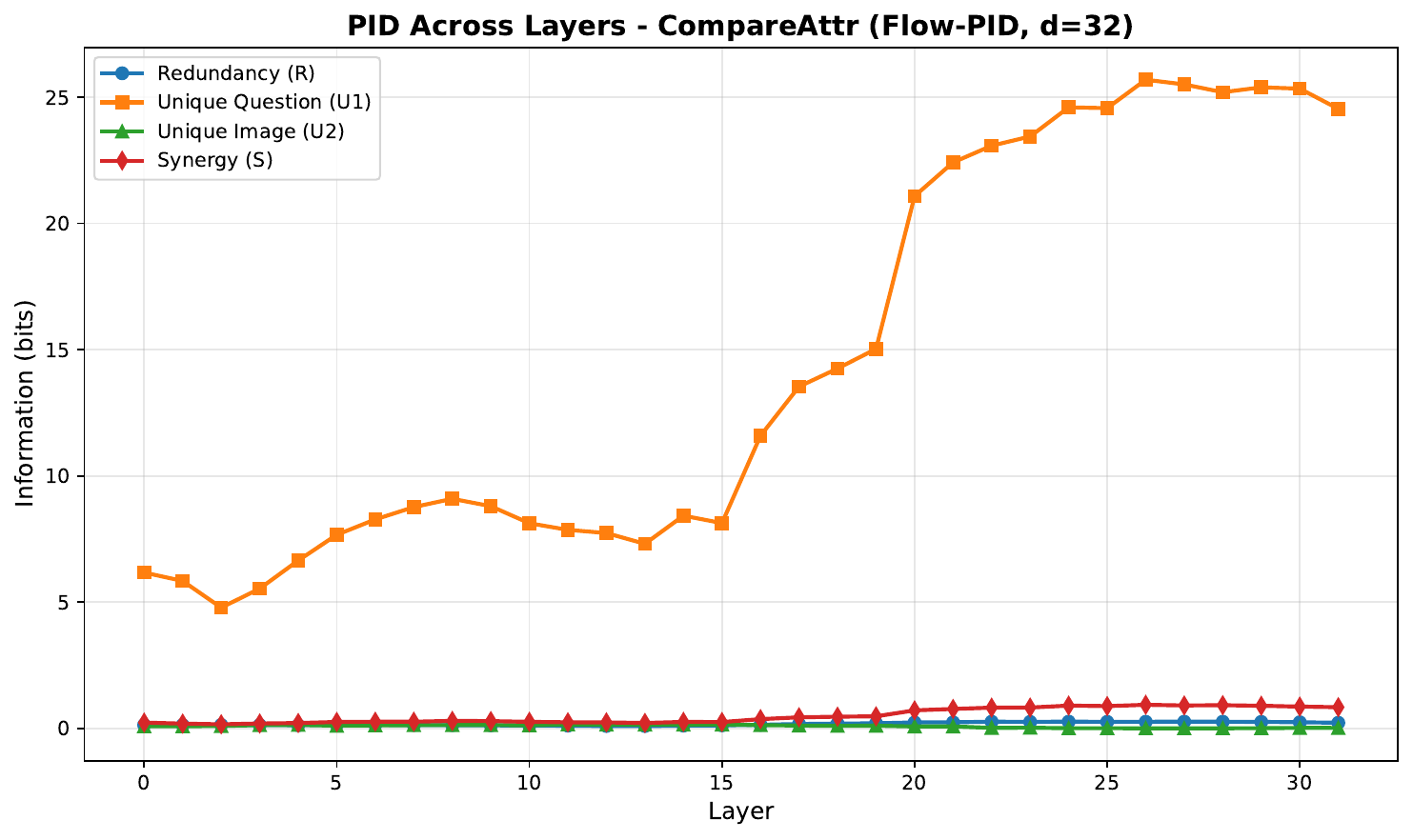}
        \caption{Compare Attribute}
    \end{subfigure}
    \hfill
    \begin{subfigure}[t]{0.32\linewidth}
        \centering
        \includegraphics[width=\linewidth]{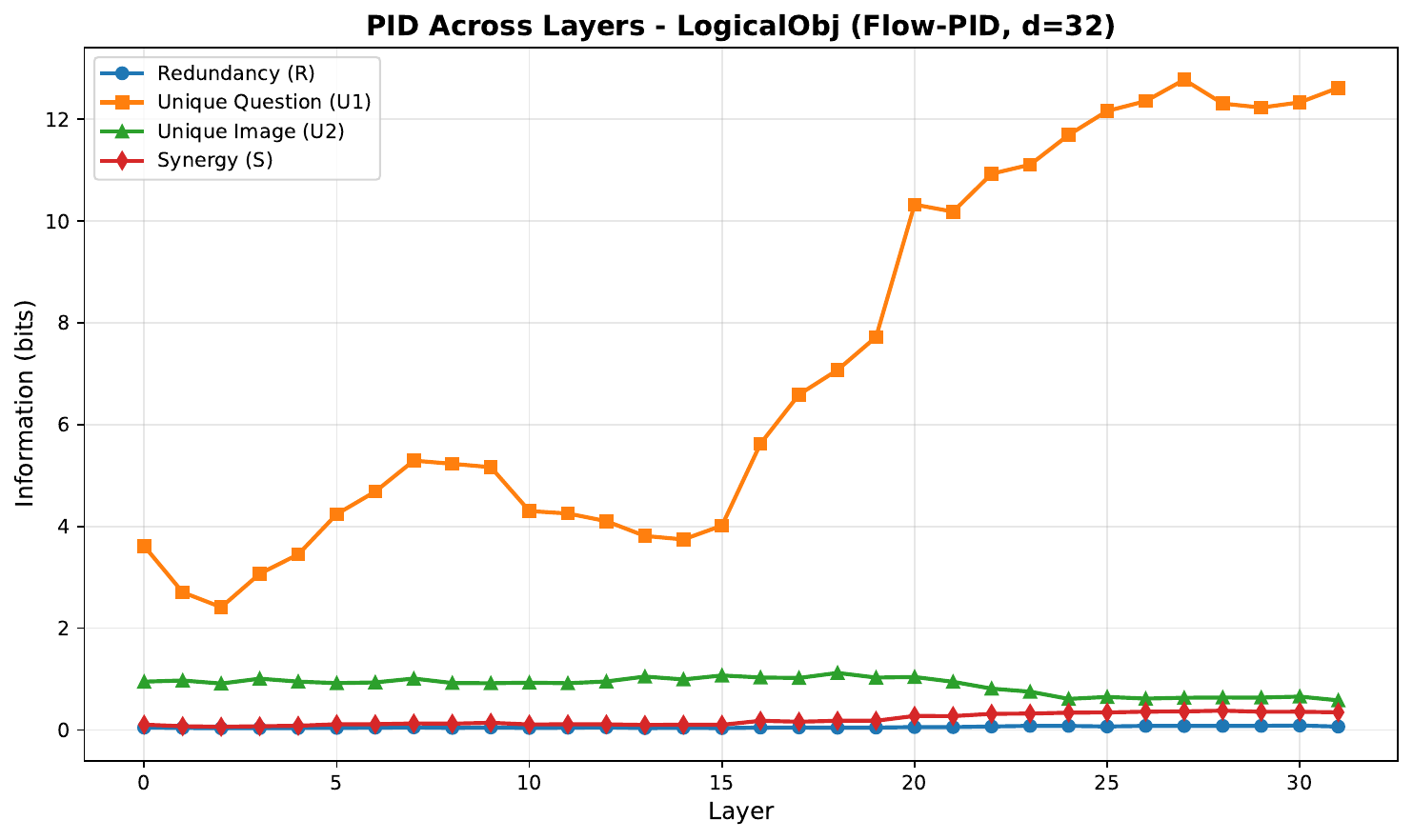}
        \caption{Logical Object}
    \end{subfigure}
    \hfill
    \begin{subfigure}[t]{0.32\linewidth}
        \centering
        \includegraphics[width=\linewidth]{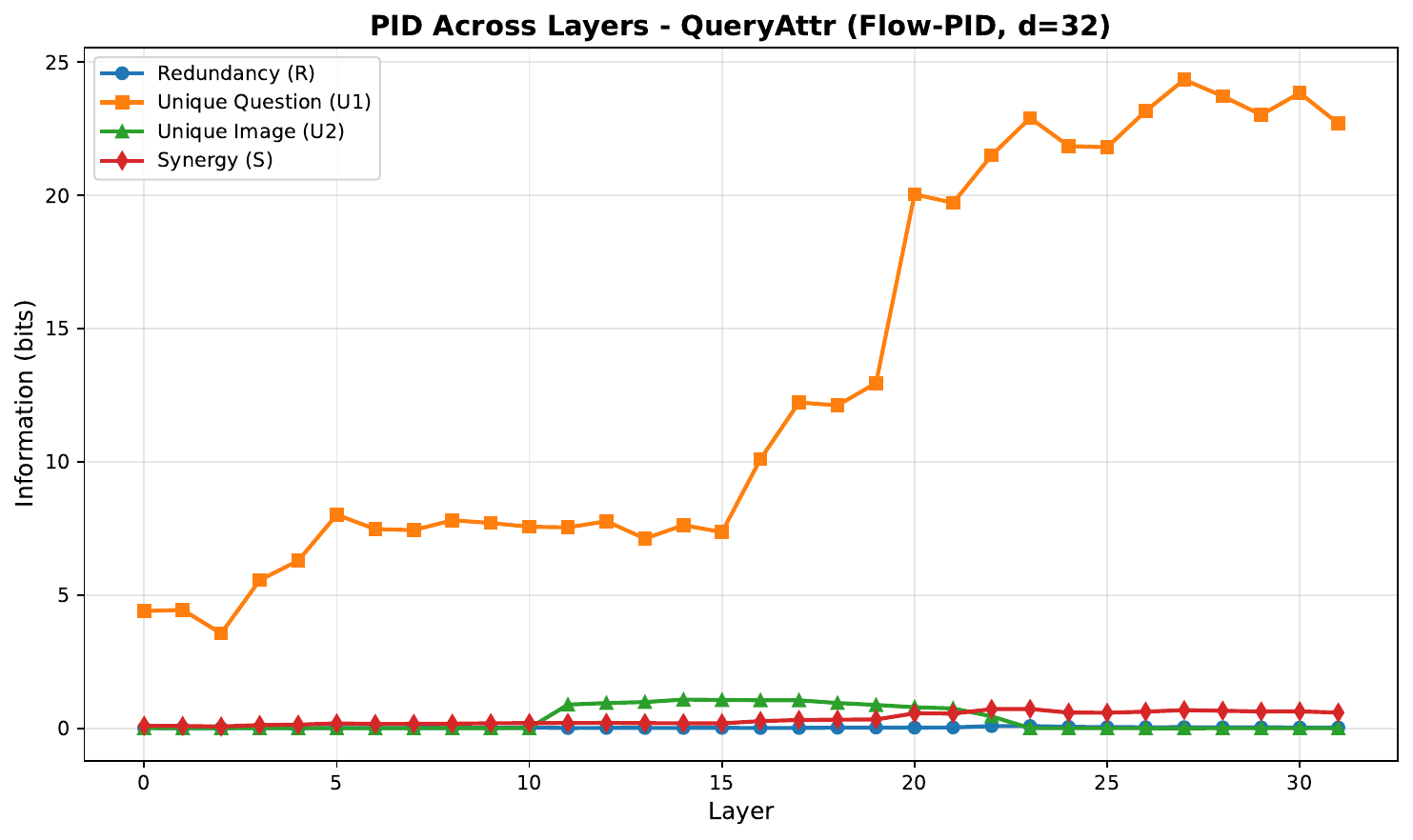}
        \caption{Query Attribute}
    \end{subfigure}

    \caption{%
    \textbf{Layer-wise PID trajectories for \textsc{LLaVA-1.5-7B} across six tasks.}
    Each panel plots redundancy~($R^\ell$), vision-unique~($U_V^\ell$),
    language-unique~($U_L^\ell$), and synergy~($S^\ell$) as functions of
    Transformer depth.  All tasks share a consistent pattern: $U_V$ peaks
    early and decays, $U_L$ surges in late layers, and $S$ remains bounded
    throughout---the signature of \emph{modal transduction}.}
    \label{fig:pid_llava_all_tasks}
\end{figure*}

Figure~\ref{fig:pid_llava_all_tasks} presents full 32-layer PID trajectories
for \textsc{LLaVA-1.5-7B} on six semantically diverse tasks: attribute
recognition (\textsc{ChooseAttr}), category selection (\textsc{ChooseCat}),
spatial relation reasoning (\textsc{ChooseRel}), attribute comparison
(\textsc{CompareAttr}), logical selection (\textsc{LogicalObj}), and
localization queries (\textsc{QueryAttr}).  Despite substantial differences in
task semantics, the trajectories share a strikingly consistent depth-wise
structure, pointing to a shared computational mechanism.

\paragraph{Three recurrent signatures identify modal transduction.}
Across all six tasks, the PID trajectories exhibit three depth-wise regularities
that collectively satisfy the modal-transduction criteria defined in
Section~\ref{sec:three_mechanisms}.
\emph{(i)~Visual-unique information decays with depth.}
$U_V$ decreases from moderate early-layer values (2--8~bits) to low final-layer
values ($<$3~bits), indicating that visual evidence is progressively consumed
and re-encoded rather than preserved as an independent stream.
\emph{(ii)~Language-unique information dominates late layers.}
$U_L$ follows a characteristic U-shaped trajectory---a mild early rise, a
mid-layer plateau or trough (layers~10--15), and a sharp late-layer surge
(layers~20--32) that ultimately controls the decision.
\emph{(iii)~Synergy remains bounded.}
$S$ stays below 3.5~bits and under 15\% of total predictive information at
every layer, inconsistent with a persistent-synergy regime.
These signatures align with the theoretical predictions of
Section~\ref{sec:three_mechanisms}: early visual injection, mid-layer
consolidation, and late language-space decision formation.

\paragraph{Final-layer composition confirms language dominance.}
Table~\ref{tab:final_layer_pid} quantifies the endpoint.  Averaged across
tasks, $U_L$ accounts for \textbf{82.4\%} of total predictive information at
Layer~31, while $U_V$ contributes only \textbf{6.4\%}
($U_L/U_V \approx 18{:}1$).  The dominance is consistent but
task-modulated.  For the high-redundancy task \textsc{ChooseCat}, $U_L$
accounts for 48.1\% of the total, with substantial redundancy
($R = 7.74$~bits) absorbing the remainder.  In contrast,
\textsc{ChooseRel}, \textsc{CompareAttr}, and \textsc{QueryAttr} each
exceed 93\% $U_L$ share, with $U_V$ approaching zero.  Synergy remains
small even in intuitively cross-modal tasks ($S = 0.65$~bits for
\textsc{ChooseRel}), providing quantitative evidence against persistent
fusion at decision time.

\begin{table}[t]
\centering
\caption{Final-layer (Layer~31) PID decomposition for \textsc{LLaVA-1.5-7B}.}
\label{tab:final_layer_pid}
\small
\setlength{\tabcolsep}{4pt}
\begin{tabularx}{\linewidth}{l
    >{\centering\arraybackslash}X
    >{\centering\arraybackslash}X
    >{\centering\arraybackslash}X
    >{\centering\arraybackslash}X
    >{\centering\arraybackslash}X}
\toprule
Task
& $R$
& $U_L$
& $U_V$
& $S$
& $U_L/\text{Total}$ \\
\midrule
ChooseAttr   & 1.13 & 14.18 & 2.01 & 1.71 & 74.5\% \\
ChooseCat    & 7.74 & 15.40 & 5.31 & 3.54 & 48.1\% \\
ChooseRel    & 0.13 & 18.83 & 0.50 & 0.65 & 93.6\% \\
CompareAttr  & 0.22 & 24.54 & 0.01 & 0.83 & 95.9\% \\
LogicalObj   & 0.07 & 12.62 & 0.58 & 0.35 & 92.7\% \\
QueryAttr    & 0.03 & 22.70 & 0.00 & 0.59 & 97.3\% \\
\midrule
\textbf{Avg.}
& \textbf{1.55}
& \textbf{18.05}
& \textbf{1.40}
& \textbf{1.28}
& \textbf{82.4\%} \\
\bottomrule
\end{tabularx}
\end{table}

\begin{figure*}[t]
  \centering
  \includegraphics[width=\textwidth]{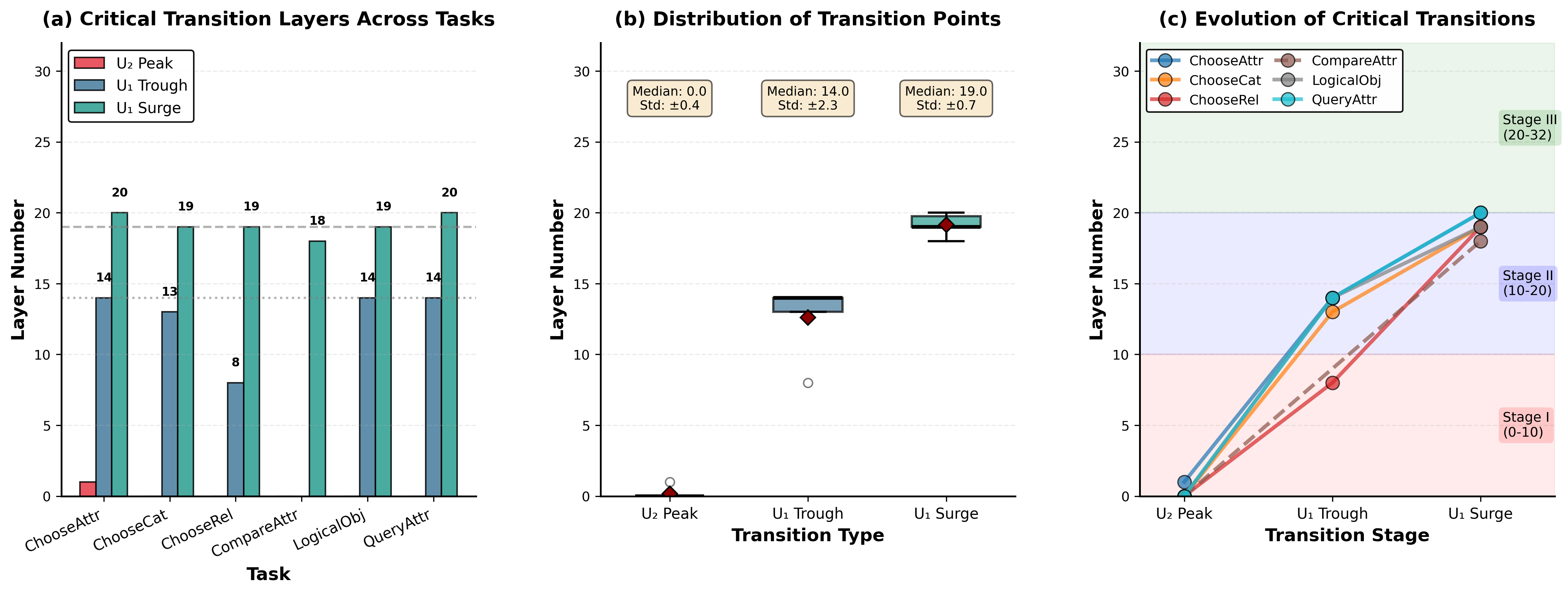}
  \caption{%
  \textbf{Turning layers and final-layer composition.}
  Transition points ($U_V$~peak, $U_L$~trough, $U_L$~surge onset)
  cluster across tasks, and final-layer predictive information is dominated
  by~$U_L$.}
  \label{fig:transitions_and_final}
\end{figure*}

\paragraph{A three-stage computation emerges from the trajectories.}
The depth-wise dynamics reveal a stable three-stage structure shared across
tasks (Figure~\ref{fig:transitions_and_final};
Table~\ref{tab:turning_points}).

\emph{Stage~I\,(Layers~0--10): visual injection.}\;
$U_V$ drops rapidly while $U_L$ increases modestly, consistent with early
cross-attention injecting visual evidence into question tokens and initiating
re-encoding in language space.

\emph{Stage~II\,(Layers~10--20): consolidation.}\;
$U_V$ continues to decline more slowly; $U_L$ temporarily stabilises or
dips; redundancy~$R$ typically increases, reflecting growing shared semantics.

\emph{Stage~III\,(Layers~20--32): language-dominant decision formation.}\;
$U_L$ rises sharply and $U_V$ reaches its minimum, indicating that late-layer
reasoning operates primarily in the language representation using
already-transduced visual evidence.

The turning points that demarcate these stages cluster tightly across tasks
(Table~\ref{tab:turning_points}): the $U_V$ peak occurs at layers~0--1, the
$U_L$ trough near layer~13, and the $U_L$ surge onset near layer~19.  This
clustering suggests that the stage structure is anchored by architectural depth
rather than task-specific idiosyncrasies.

\begin{table}[t]
\centering
\caption{Cross-task consistency of modal-transduction turning points.}
\label{tab:turning_points}
\small
\setlength{\tabcolsep}{6pt}
\begin{tabularx}{\linewidth}{l
    >{\centering\arraybackslash}X
    >{\centering\arraybackslash}X
    >{\centering\arraybackslash}X}
\toprule
Task
& Peak of $U_V$
& Trough of $U_L$
& Onset of $U_L$ surge \\
\midrule
ChooseAttr   & 0  & 13 & 20 \\
ChooseCat    & 1  & 14 & 19 \\
ChooseRel    & 0  & 11 & 19 \\
CompareAttr  & 1  & -- & 17 \\
LogicalObj   & 0  & 13 & 19 \\
QueryAttr    & 1  & 13 & 20 \\
\midrule
\textbf{Median}
& \textbf{0.5}
& \textbf{13}
& \textbf{19} \\
\bottomrule
\end{tabularx}
\end{table}

\paragraph{Task semantics modulate information fingerprints without changing
the mechanism.}
While modal transduction is universal across the six tasks, task semantics
shape the detailed information composition---what we term the task's
\emph{information fingerprint}.
\textsc{ChooseCat} exhibits the highest redundancy ($R = 7.74$~bits, 24.2\%
share), consistent with category concepts being strongly encoded in both
modalities.
Relational, comparative, and localization tasks (\textsc{ChooseRel},
\textsc{CompareAttr}, \textsc{QueryAttr}) instead show near-zero redundancy
($R < 0.3$~bits) and very high $U_L$ shares ($>$93\%), indicating a pattern of
\emph{language reasoning under visual conditioning}: vision supplies
entity- and attribute-level evidence early, while the decisive operations
unfold in language space.
The extreme case, \textsc{QueryAttr}, has $U_V \approx 0$ at the final
layer, consistent with multi-step reasoning proceeding almost entirely within
language representations after early extraction.

\paragraph{An information-theoretic reinterpretation of attention-knockout
findings.}
The PID trajectories also offer a new lens on prior attention-knockout
results.  A pathway can have a small \emph{late-stage} knockout effect not
because vision is irrelevant, but because decision-relevant visual evidence
has already been transduced into question representations at earlier layers.
In this view, early-layer Image$\rightarrow$Question connections are causal
bottlenecks for the $U_V \!\to\! U_L$ transfer, whereas late-stage
Question$\rightarrow$Last connections primarily convey the amplified $U_L$
signal that dominates readout.
We validate this interpretation with targeted interventions in
Section~\ref{subsec:results_knockout}.

\begin{figure*}[t]
  \centering
  \includegraphics[width=\textwidth]{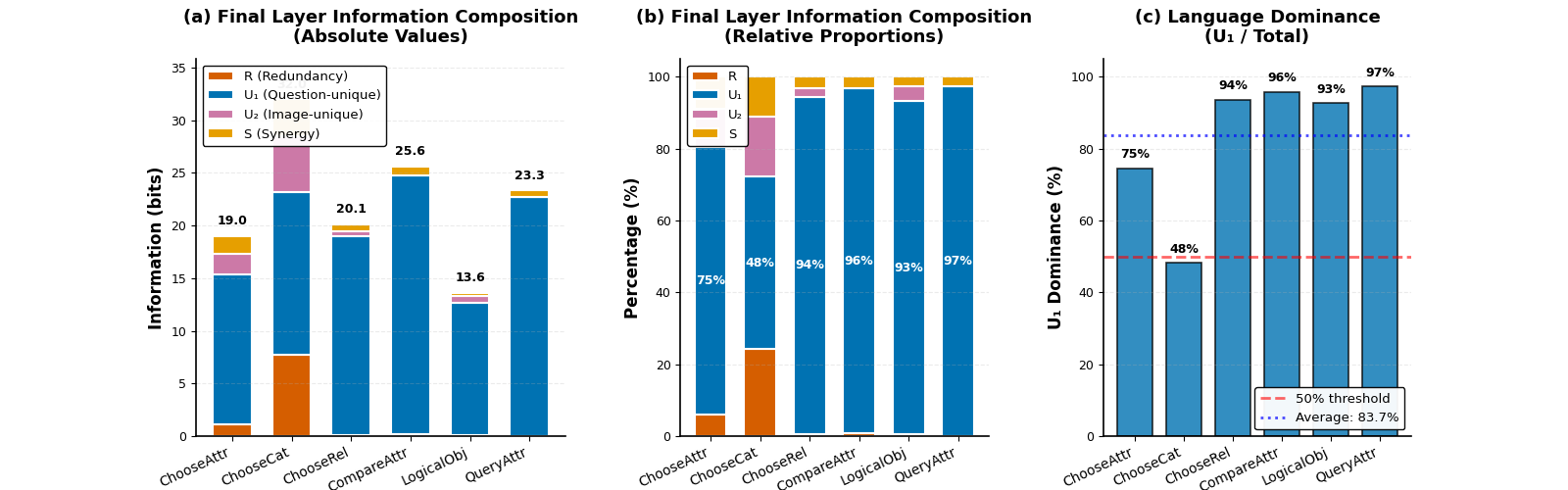}
  \caption{%
  \textbf{Layer-wise PID trajectories for \textsc{LLaVA-1.5-7B} across six
  tasks.}  All tasks exhibit decay of $U_V$, a late-stage surge of $U_L$,
  and bounded synergy~$S$, supporting modal transduction as the dominant
  mechanism.}
  \label{fig:pid_traj_main}
\end{figure*}

\subsection{Results II: Cross-Model Consistency}
\label{subsec:results_cross_model}

Section~\ref{subsec:results_modal_transduction} establishes modal transduction
in \textsc{LLaVA-1.5-7B}.  To test whether these patterns generalise beyond a
single implementation, we replicate the full pipeline on \textsc{LLaVA-1.6-7B}.
The comparison is non-trivial: \textsc{LLaVA-1.6} introduces dynamic
high-resolution visual processing with variable image-token counts (up to
$4\times$ those of \textsc{LLaVA-1.5}) and a different Vicuna-family language
backbone and training mix, while sharing the same CLIP ViT-L vision encoder and
32-layer Transformer depth.  If modal transduction reflects a deeper organising
principle---pretrained LLM priors, task semantics, or depth-induced
computation---then three properties should be stable across models:
(i)~final-layer PID composition, (ii)~full layer-wise trajectories, and
(iii)~key turning points.  We test each in turn.

\begin{figure*}[t]
    \centering
    \begin{subfigure}[t]{0.32\linewidth}
        \centering
        \includegraphics[width=\linewidth]{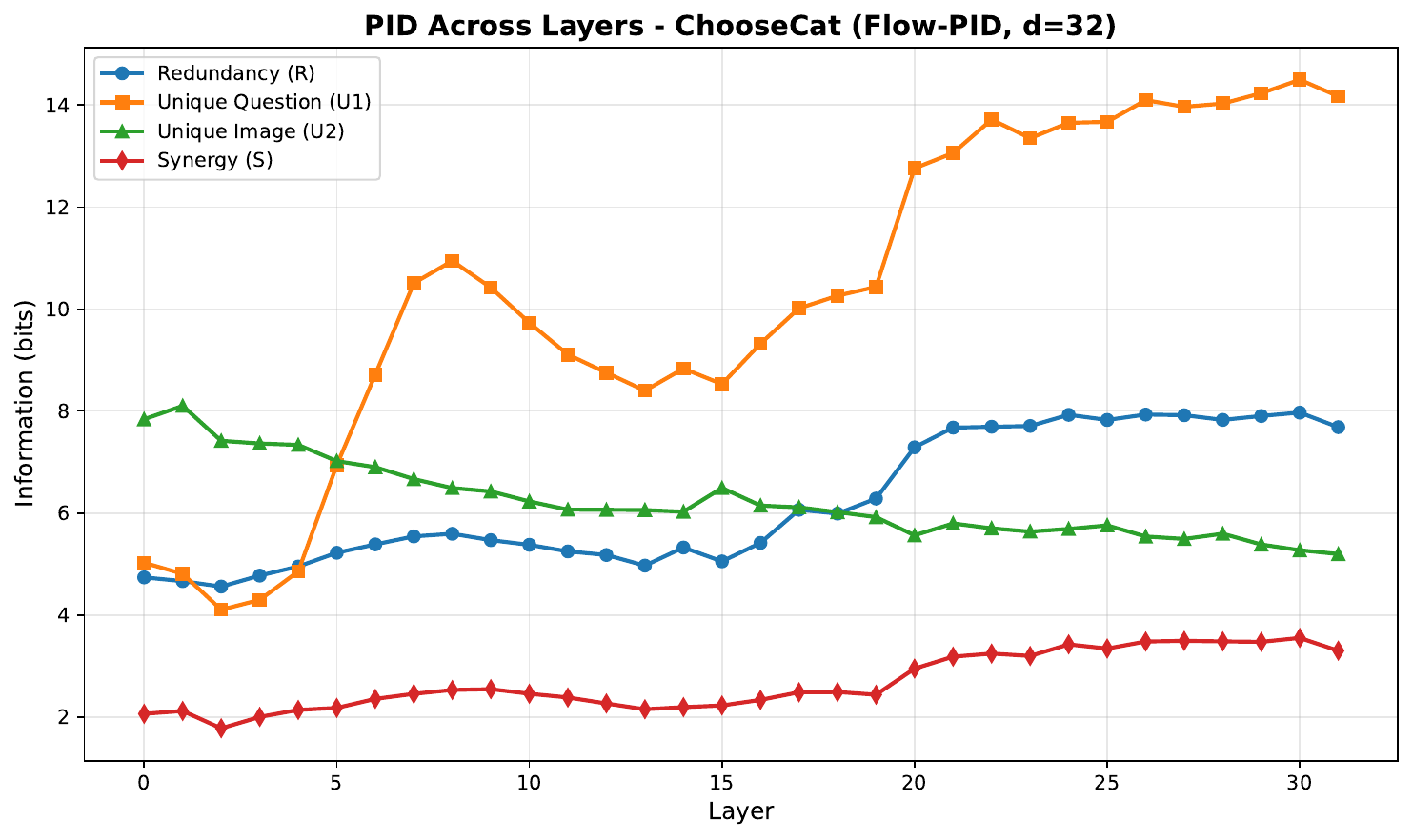}
        \caption{Choose Attribute}
    \end{subfigure}
    \hfill
    \begin{subfigure}[t]{0.32\linewidth}
        \centering
        \includegraphics[width=\linewidth]{image/ChooseCat.pdf}
        \caption{Choose Category}
    \end{subfigure}
    \hfill
    \begin{subfigure}[t]{0.32\linewidth}
        \centering
        \includegraphics[width=\linewidth]{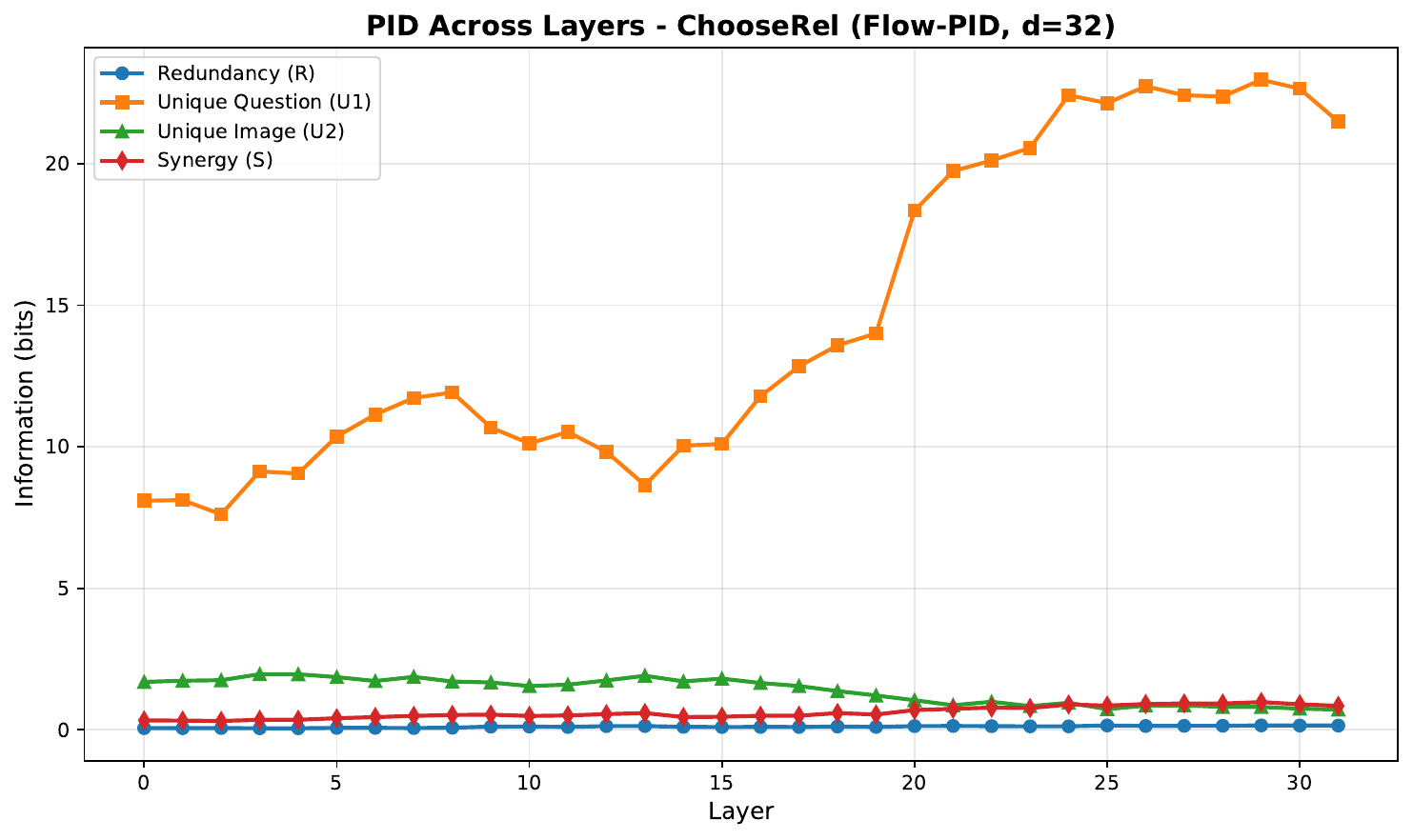}
        \caption{Choose Relation}
    \end{subfigure}

    \begin{subfigure}[t]{0.32\linewidth}
        \centering
        \includegraphics[width=\linewidth]{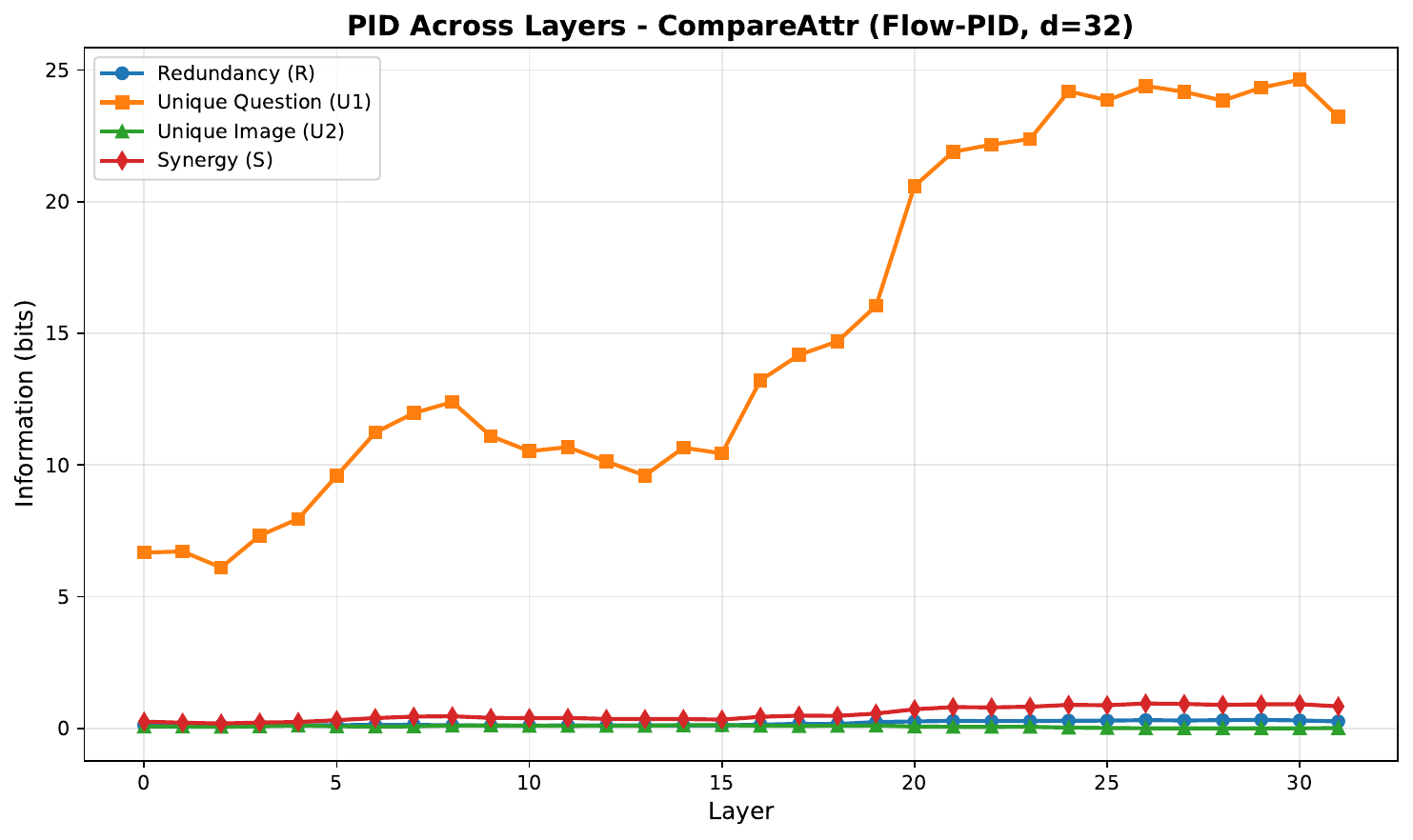}
        \caption{Compare Attribute}
    \end{subfigure}
    \hfill
    \begin{subfigure}[t]{0.32\linewidth}
        \centering
        \includegraphics[width=\linewidth]{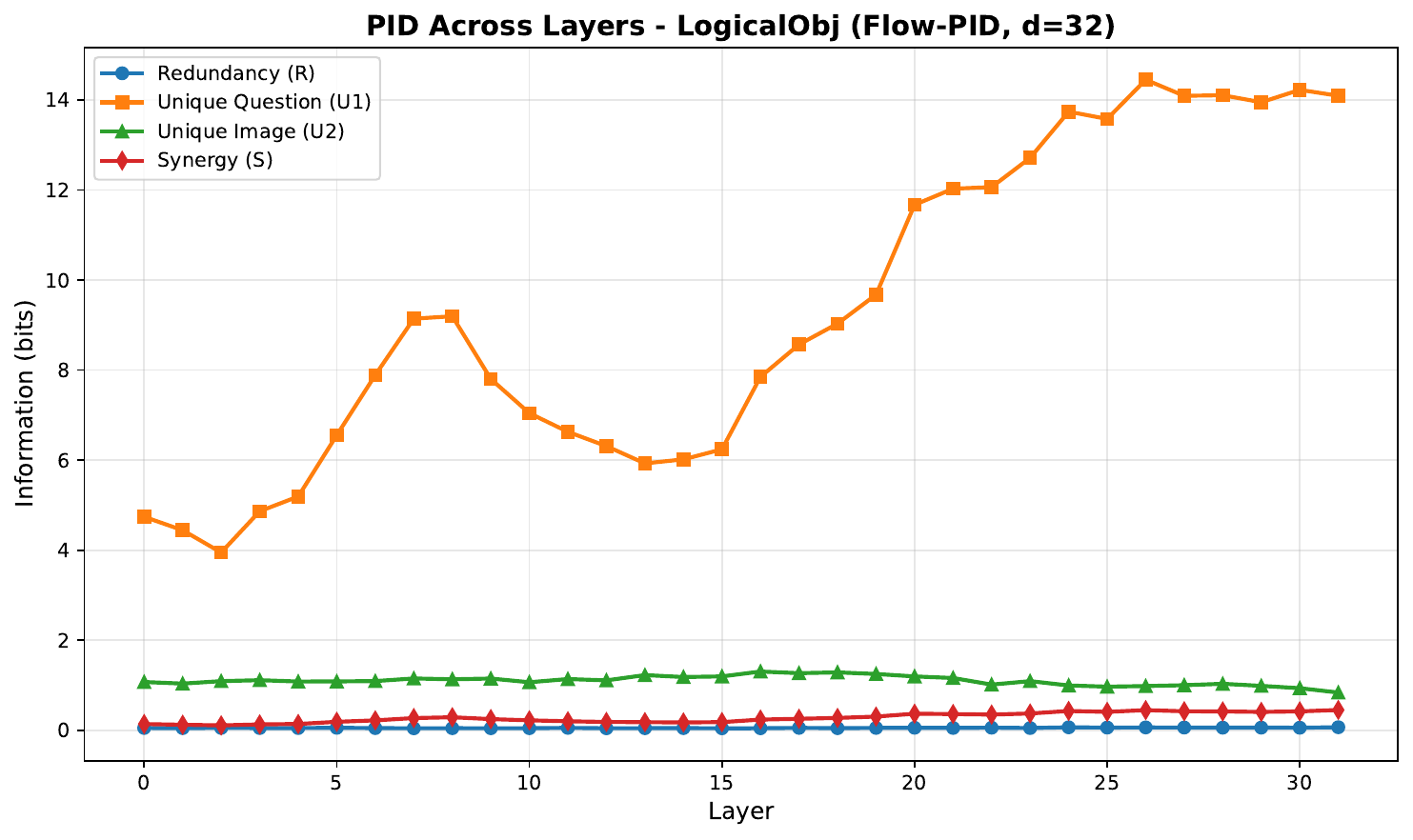}
        \caption{Logical Object}
    \end{subfigure}
    \hfill
    \begin{subfigure}[t]{0.32\linewidth}
        \centering
        \includegraphics[width=\linewidth]{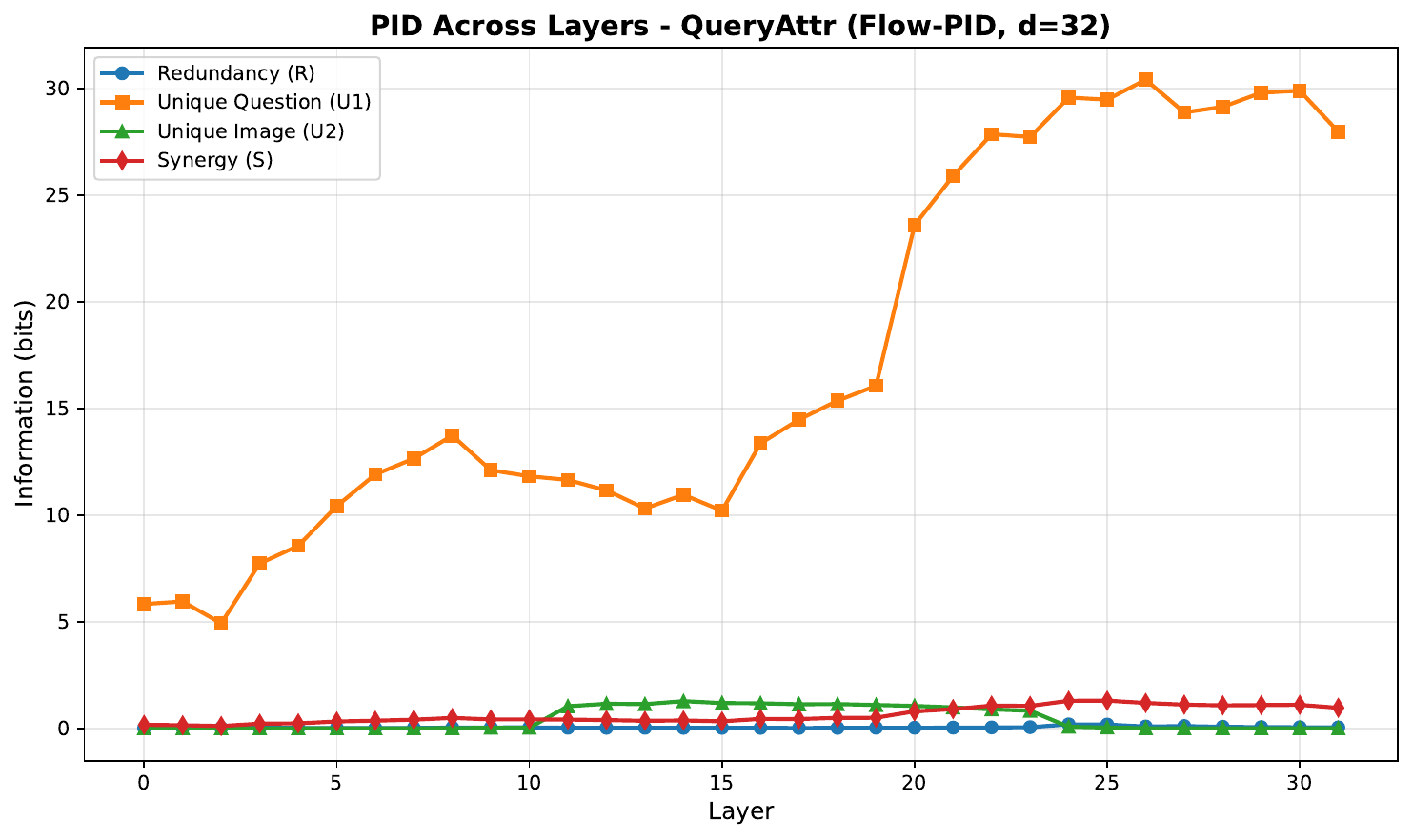}
        \caption{Query Attribute}
    \end{subfigure}

    \caption{%
    \textbf{Layer-wise PID trajectories for \textsc{LLaVA-1.6-7B} across six
    tasks.}  The depth-wise pattern mirrors \textsc{LLaVA-1.5-7B}
    (Figure~\ref{fig:pid_llava_all_tasks}): $U_V$ decays, $U_L$ surges late,
    and synergy remains bounded, confirming modal transduction in an
    architecturally distinct model.}
    \label{fig:pid_all_tasks}
\end{figure*}

\paragraph{Final-layer invariance: language dominance persists.}

\begin{figure*}[t]
\centering
\includegraphics[width=\linewidth]{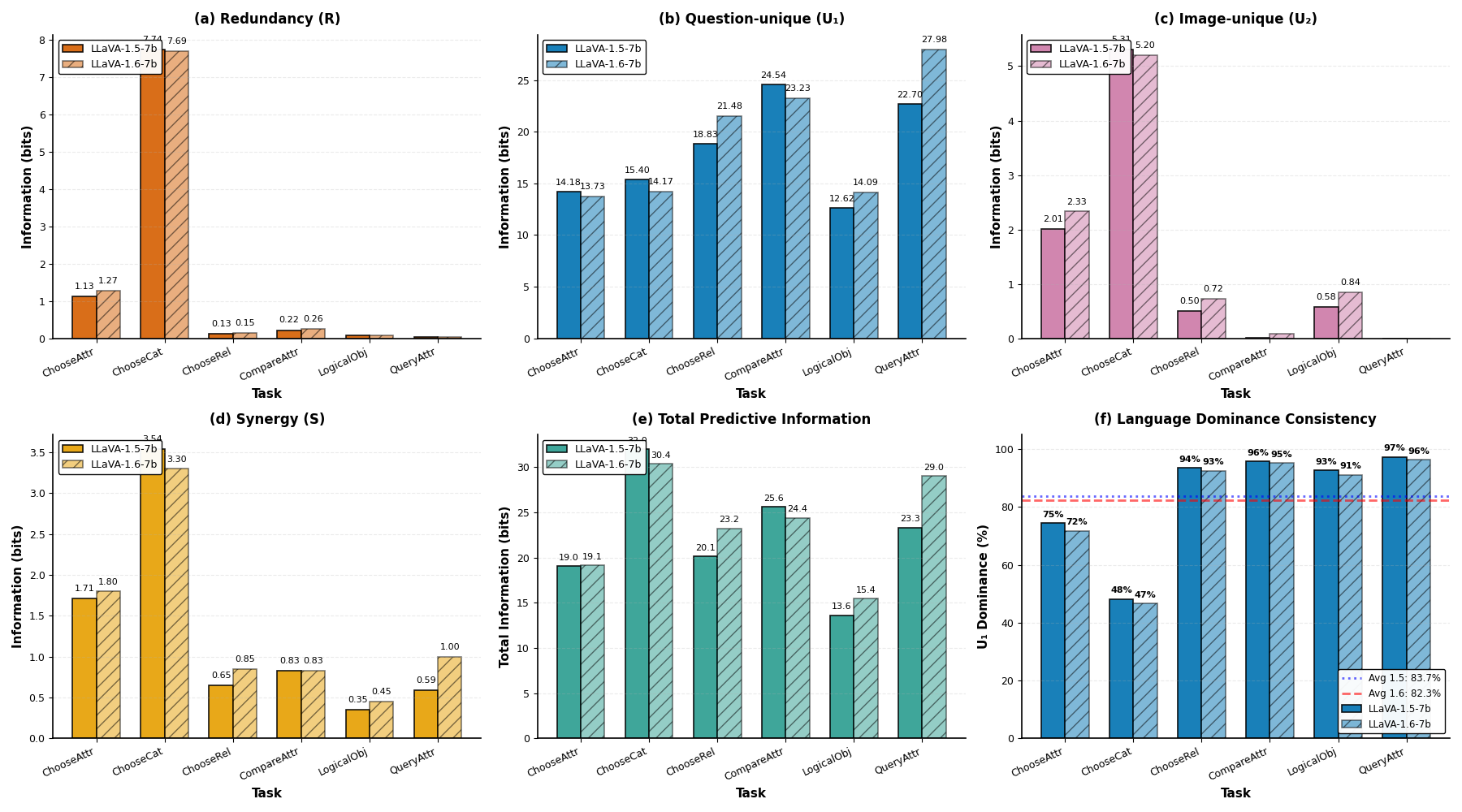}
\caption{%
\textbf{Final-layer PID composition across models.}
$U_L$ dominates at Layer~31 in both \textsc{LLaVA-1.5-7B} and
\textsc{LLaVA-1.6-7B}, while synergy remains bounded.}
\label{fig:cross_model_final_layer}
\end{figure*}

Table~\ref{tab:final_layer_pid_cross_model} compares the Layer~31 PID
decomposition between the two models across all six tasks
(Figure~\ref{fig:cross_model_final_layer} provides a visual summary).
Two conclusions emerge.

First, \emph{language dominance is nearly unchanged}.  Averaged across tasks,
the $U_L$ share is \textbf{82.0\%} for \textsc{LLaVA-1.5-7B} and
\textbf{80.7\%} for \textsc{LLaVA-1.6-7B}---a difference of only 1.3
percentage points.  Task-level variation is similarly small (e.g.,
\textsc{ChooseAttr}: 74.5\% vs.\ 71.8\%; \textsc{ChooseRel}: 93.6\% vs.\
92.6\%), and synergy stays bounded ($S < 4$~bits, $< 15\%$ of total) in all
cases.

Second, \emph{task-level information fingerprints are preserved}.
\textsc{ChooseCat} remains the canonical high-redundancy case in both models
($R \approx 7.7$--$7.8$~bits; lowest $U_L$ share $\approx 47$--$48\%$),
while relational, comparative, and localization tasks retain strong language
dominance ($U_L > 90\%$) and near-zero redundancy ($R < 0.3$~bits).  Where
models do differ, the change is in \emph{magnitude} rather than
\emph{organisation}: \textsc{LLaVA-1.6} shows higher total information for
some tasks (e.g., \textsc{ChooseRel}: $+15\%$; \textsc{QueryAttr}: $+24\%$),
consistent with improved visual resolution increasing the \emph{amount} of
information available for transduction without altering the transduction
mechanism itself.

\begin{table*}[t]
\centering
\caption{\textsc{LLaVA-1.5-7B} vs.\ \textsc{LLaVA-1.6-7B}: final-layer (Layer~31) PID comparison.}
\label{tab:final_layer_pid_cross_model}
\small
\setlength{\tabcolsep}{4pt}
\renewcommand{\arraystretch}{0.95}
\resizebox{\textwidth}{!}{%
\begin{tabular}{llrrrrrrrr}
\toprule
Task & Model & $R$ & $U_L$ & $U_V$ & $S$ & Total & $U_L$ Share & $U_L/U_V$ \\
\midrule
\textbf{ChooseAttr} & 1.5-7b & 1.13 & 14.18 & 2.01 & 1.71 & 19.03 & 74.5\% & $7.05\times$ \\
                   & 1.6-7b & 1.27 & 13.73 & 2.33 & 1.80 & 19.13 & 71.8\% & $5.89\times$ \\
                   & \textbf{Diff.} & \textbf{+12\%} & \textbf{-3.2\%} & \textbf{+16\%} & \textbf{+5.3\%} & \textbf{+0.5\%} & \textbf{-2.7\%} & \textbf{-16\%} \\
\midrule
\textbf{ChooseCat}  & 1.5-7b & 7.74 & 15.40 & 5.31 & 3.54 & 31.99 & 48.1\% & $2.90\times$ \\
                   & 1.6-7b & 7.69 & 14.17 & 5.20 & 3.30 & 30.36 & 46.7\% & $2.73\times$ \\
                   & \textbf{Diff.} & \textbf{-0.6\%} & \textbf{-8.0\%} & \textbf{-2.1\%} & \textbf{-6.8\%} & \textbf{-5.1\%} & \textbf{-1.4\%} & \textbf{-5.9\%} \\
\midrule
\textbf{ChooseRel}  & 1.5-7b & 0.13 & 18.83 & 0.50 & 0.65 & 20.11 & 93.6\% & $37.66\times$ \\
                   & 1.6-7b & 0.15 & 21.48 & 0.72 & 0.85 & 23.20 & 92.6\% & $29.83\times$ \\
                   & \textbf{Diff.} & \textbf{+15\%} & \textbf{+14\%} & \textbf{+44\%} & \textbf{+31\%} & \textbf{+15\%} & \textbf{-1.0\%} & \textbf{-21\%} \\
\midrule
\textbf{CompareAttr}& 1.5-7b & 0.22 & 24.54 & 0.01 & 0.83 & 25.60 & 95.9\% & $2454\times$ \\
                   & 1.6-7b & 0.26 & 23.23 & 0.08 & 0.83 & 24.40 & 95.2\% & $290\times$ \\
                   & \textbf{Diff.} & \textbf{+18\%} & \textbf{-5.3\%} & \textbf{+700\%} & \textbf{0\%} & \textbf{-4.7\%} & \textbf{-0.7\%} & \textbf{-88\%} \\
\midrule
\textbf{LogicalObj} & 1.5-7b & 0.07 & 12.62 & 0.58 & 0.35 & 13.62 & 92.7\% & $21.76\times$ \\
                   & 1.6-7b & 0.07 & 14.09 & 0.84 & 0.45 & 15.45 & 91.2\% & $16.77\times$ \\
                   & \textbf{Diff.} & \textbf{0\%} & \textbf{+12\%} & \textbf{+45\%} & \textbf{+29\%} & \textbf{+13\%} & \textbf{-1.5\%} & \textbf{-23\%} \\
\midrule
\textbf{QueryAttr}  & 1.5-7b & 0.03 & 22.70 & 0.00 & 0.59 & 23.32 & 97.3\% & $\infty$ \\
                   & 1.6-7b & 0.04 & 27.98 & 0.00 & 0.96 & 28.98 & 96.6\% & $\infty$ \\
                   & \textbf{Diff.} & \textbf{+33\%} & \textbf{+23\%} & \textbf{--} & \textbf{+63\%} & \textbf{+24\%} & \textbf{-0.7\%} & \textbf{--} \\
\midrule
\textbf{Avg.}       & 1.5-7b & 1.55 & 18.05 & 1.40 & 1.28 & 22.28 & \textbf{82.0\%} & \textbf{$\sim 18\times$} \\
                   & 1.6-7b & 1.58 & 19.11 & 1.53 & 1.37 & 23.59 & \textbf{80.7\%} & \textbf{$\sim 11\times$} \\
                   & \textbf{Avg.\ Diff.} & \textbf{+2\%} & \textbf{+5.9\%} & \textbf{+9.3\%} & \textbf{+7.0\%} & \textbf{+5.9\%} & \textbf{-1.3\%} & \textbf{--} \\
\bottomrule
\end{tabular}}
\end{table*}

\paragraph{Trajectory-level invariance: depth-wise dynamics are synchronised.}

\begin{figure*}[t]
\centering
\includegraphics[width=\linewidth]{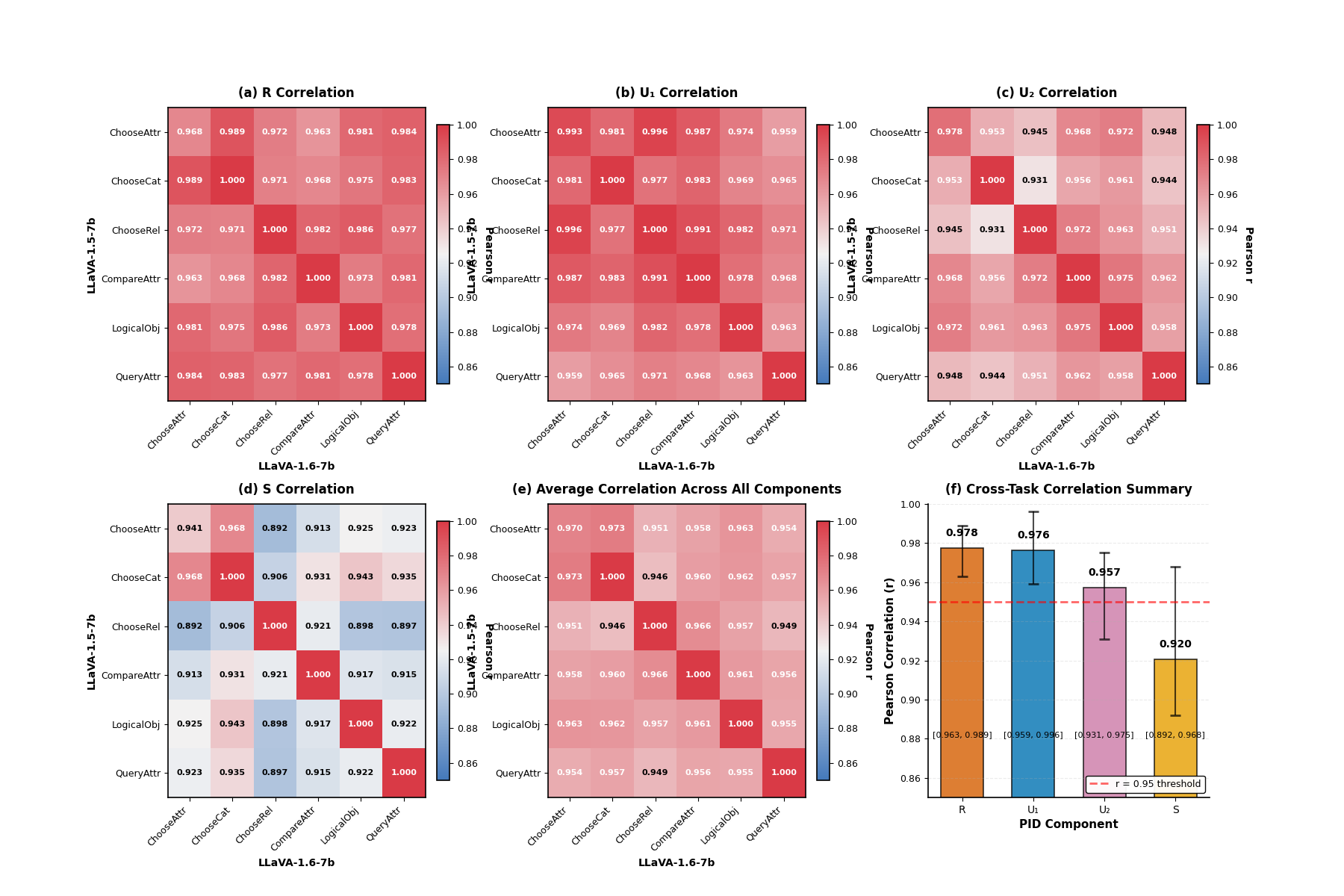}
\caption{%
\textbf{Cross-model trajectory alignment.}
Solid: \textsc{LLaVA-1.5-7B}; dashed: \textsc{LLaVA-1.6-7B}.
Depth-wise dynamics of all four PID components are closely synchronised.}
\label{fig:cross_model_trajectory}
\end{figure*}

Final-layer agreement does not guarantee mechanistic agreement across depth.
Figure~\ref{fig:cross_model_trajectory} overlays representative PID
trajectories from both models, showing closely aligned depth-wise evolution:
$U_V$ decays monotonically, $U_L$ follows the same U-shaped profile with a
late surge, and $S$ remains low throughout.

Table~\ref{tab:trajectory_correlation} quantifies this alignment via
cross-model Pearson correlations computed across the 32 layers for each PID
component.  All components are highly correlated (mean $r$:
$U_L = 0.982$, $R = 0.976$, $U_V = 0.961$, $S = 0.927$), confirming that
the two models match not only in endpoint composition but also in their
\emph{depth-wise information dynamics}.  Synergy exhibits slightly lower
correlation, consistent with higher estimation variance for synergistic terms
in high-dimensional settings, but its qualitative behaviour---bounded and
non-dominant---is invariant across models.

\begin{table}[t]
\centering
\caption{Cross-model PID trajectory correlation (Pearson $r$ across 32 layers).}
\label{tab:trajectory_correlation}
\small
\setlength{\tabcolsep}{6pt}
\renewcommand{\arraystretch}{1.0}
\begin{tabular}{lrrrrr}
\toprule
Task & $r(R)$ & $r(U_L)$ & $r(U_V)$ & $r(S)$ & Mean \\
\midrule
ChooseAttr  & 0.968 & 0.993 & 0.978 & 0.941 & 0.970 \\
ChooseCat   & 0.989 & 0.981 & 0.953 & 0.968 & 0.973 \\
ChooseRel   & 0.972 & 0.996 & 0.945 & 0.892 & 0.951 \\
CompareAttr & 0.963 & 0.987 & 0.968 & 0.913 & 0.958 \\
LogicalObj  & 0.981 & 0.974 & 0.972 & 0.925 & 0.963 \\
QueryAttr   & 0.984 & 0.959 & 0.948 & 0.923 & 0.954 \\
\midrule
\textbf{Avg.} & \textbf{0.976} & \textbf{0.982} & \textbf{0.961} & \textbf{0.927} & \textbf{0.962} \\
\bottomrule
\end{tabular}
\end{table}

\paragraph{Turning-point stability: stage transitions are depth-anchored.}

Table~\ref{tab:turning_points_cross_model} compares three stage-transition
landmarks across models: (i)~the peak layer of $U_V$, (ii)~the trough of
$U_L$, and (iii)~the onset of the late $U_L$ surge.  The median surge onset
differs by only one layer (Layer~19 vs.\ 18), with cross-task dispersion of
$\pm 1.1$ layers.  This stability indicates that stage transitions are anchored
by architectural depth rather than visual-tokenisation details.  The sole
notable exception is \textsc{QueryAttr}, where the $U_V$ peak is delayed by
four layers in \textsc{LLaVA-1.6}; crucially, the downstream trough and surge
onset remain aligned, preserving the overall three-stage organisation.

\begin{table}[t]
\centering
\caption{Cross-model comparison of key turning points.}
\label{tab:turning_points_cross_model}
\small
\setlength{\tabcolsep}{5pt}
\renewcommand{\arraystretch}{1.0}
\begin{tabular}{llrrr}
\toprule
Task & Turning point & \textsc{1.5-7B} & \textsc{1.6-7B} & Diff. \\
\midrule
\textbf{ChooseAttr}  & $U_V$ peak       & 1  & 0  & $-1$ \\
                     & $U_L$ trough     & 14 & 15 & $+1$ \\
                     & $U_L$ surge onset& 20 & 19 & $-1$ \\
\midrule
\textbf{ChooseCat}   & $U_V$ peak       & 0  & 1  & $+1$ \\
                     & $U_L$ trough     & 13 & 15 & $+2$ \\
                     & $U_L$ surge onset& 19 & 18 & $-1$ \\
\midrule
\textbf{ChooseRel}   & $U_V$ peak       & 0  & 0  & 0 \\
                     & $U_L$ trough     & 8  & -- & -- \\
                     & $U_L$ surge onset& 19 & 18 & $-1$ \\
\midrule
\textbf{CompareAttr} & $U_V$ peak       & 0  & 0  & 0 \\
                     & $U_L$ trough     & -- & -- & -- \\
                     & $U_L$ surge onset& 18 & 17 & $-1$ \\
\midrule
\textbf{LogicalObj}  & $U_V$ peak       & 0  & 0  & 0 \\
                     & $U_L$ trough     & 14 & 15 & $+1$ \\
                     & $U_L$ surge onset& 19 & 19 & 0 \\
\midrule
\textbf{QueryAttr}   & $U_V$ peak       & 0  & 4  & $+4$ \\
                     & $U_L$ trough     & 14 & 15 & $+1$ \\
                     & $U_L$ surge onset& 20 & 18 & $-2$ \\
\midrule
\textbf{Median}      & $U_V$ peak       & 0  & 0  & 0 \\
                     & $U_L$ trough     & 14 & 15 & $+1$ \\
                     & $U_L$ surge onset& 19 & 18 & $-1$ \\
\textbf{Std.\ dev.}  & --               & $\pm 0.8$ & $\pm 0.9$ & $\pm 1.1$ \\
\bottomrule
\end{tabular}
\end{table}

\medskip

In summary, modal transduction is stable across an architecturally distinct
model at all three levels of analysis: final-layer composition
(Table~\ref{tab:final_layer_pid_cross_model}), full depth-wise dynamics
(Table~\ref{tab:trajectory_correlation}), and stage-transition landmarks
(Table~\ref{tab:turning_points_cross_model}).  These results indicate that the
transduction pattern is not a model-specific artifact but a strategy shaped
jointly by pretrained LLM priors, task semantics, and depth-induced
computational stratification.

\subsection{Results III: Causal Validation via Attention Knockout}
\label{subsec:results_knockout}

Experiments~I--II establish modal transduction observationally and demonstrate
cross-model invariance, but do not yet provide direct causal evidence.  We now
perform an \emph{attention knockout} intervention---blocking a specific
candidate pathway---and test whether the resulting PID shifts match
mechanism-level predictions.

\paragraph{Intervention design.}
For \textsc{LLaVA-1.6-7B}, we set the pre-softmax attention logits on all
\emph{Image$\rightarrow$Question} edges to $-\infty$ across all 32 layers,
severing the dominant route through which visual evidence is injected into
question tokens (implementation details in
Appendix~\ref{app:knockout_design}).  If modal transduction operates as a
progressive transfer of $U_V$ into $U_L$ via this pathway, the knockout should
produce three predictable shifts at the decision layer:

\begin{quote}
\emph{(P1)}\;$\Delta U_V > 0$: visual information becomes ``trapped'' in the
image stream instead of being consumed.\\[2pt]
\emph{(P2)}\;$\Delta S > 0$: the model must resort to less efficient joint
processing, increasing synergy.\\[2pt]
\emph{(P3)}\;$\Delta\,\text{Total} > 0$: compensating for blocked transduction
requires a larger total information budget.
\end{quote}

\noindent We further expect the \emph{magnitude} of these effects to vary with
task semantics: low-redundancy, vision-dependent tasks should show strong
sensitivity, while high-redundancy tasks should be robust due to alternative
informational routes.

\subsubsection{Decision-layer effects}
\label{subsubsec:knockout_decision}

\begin{figure*}[t]
\centering
\includegraphics[width=\linewidth]{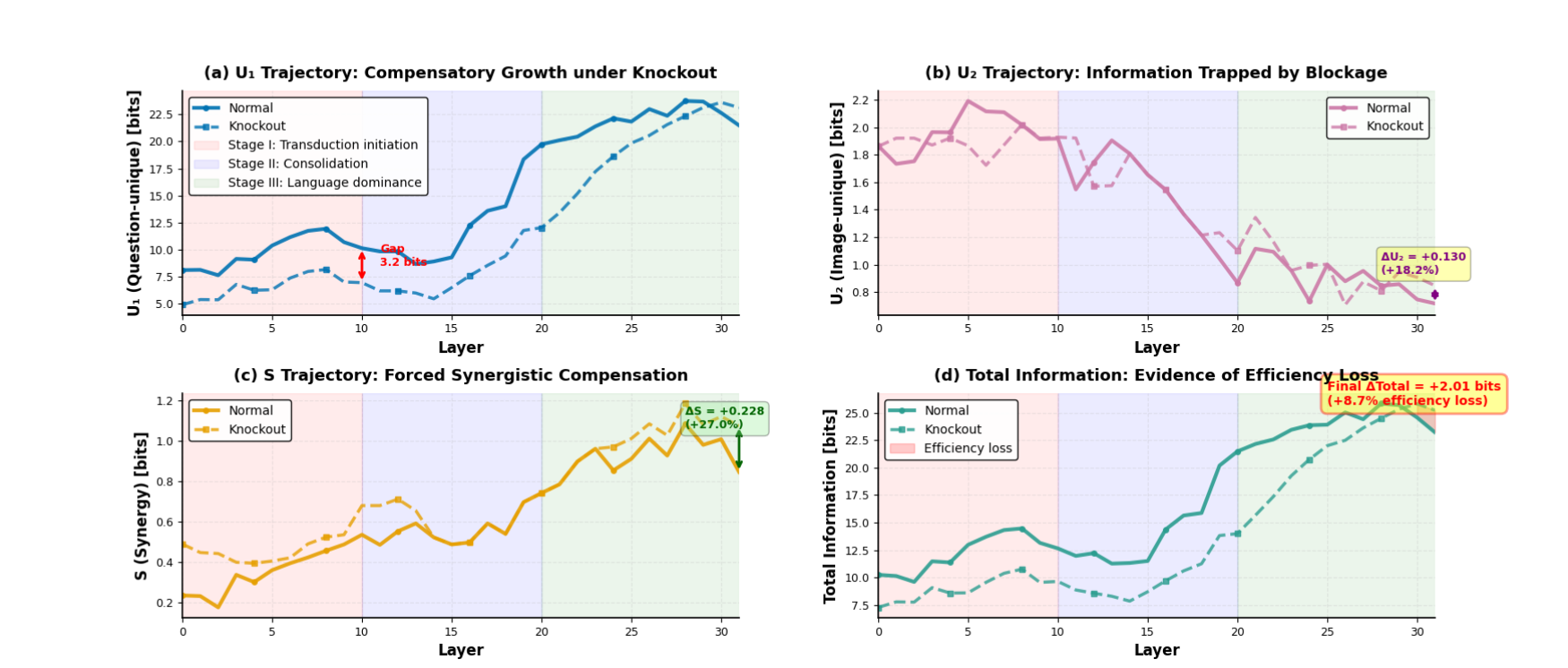}
\caption{%
\textbf{Final-layer effects of Image$\rightarrow$Question attention knockout.}
Vision-dependent tasks (\textsc{ChooseRel}, \textsc{LogicalObj}) show the
strongest PID shifts; high-redundancy tasks (\textsc{ChooseAttr},
\textsc{ChooseCat}) are largely robust.}
\label{fig:knockout_overview}
\end{figure*}

Table~\ref{tab:knockout_pid_effects} reports the Layer~31 PID composition under
normal and knockout conditions for all six tasks;
Figure~\ref{fig:knockout_overview} provides a visual summary.  The results
reveal three distinct response classes.

\begin{table*}[t]
\centering
\caption{PID effects of \emph{Image$\rightarrow$Question} attention knockout at
Layer~31 on \textsc{LLaVA-1.6-7B}.  Values are in bits;
$\Delta := (\text{knockout} - \text{normal})/\text{normal} \times 100\%$.
Bold entries mark changes exceeding $5\%$ or mechanism-critical shifts.}
\label{tab:knockout_pid_effects}
\small
\setlength{\tabcolsep}{4pt}
\renewcommand{\arraystretch}{0.95}
\resizebox{\textwidth}{!}{%
\begin{tabular}{llrrrrrrr}
\toprule
Task & Condition & $R$ & $U_L$ & $U_V$ & $S$ & Total & $U_L/U_V$ & $\Delta\,\text{Total}$ \\
\midrule
\textbf{ChooseRel} & Normal   & 0.15 & 21.48 & 0.72 & 0.85 & 23.20 & $29.8\times$ & -- \\
(High vision dep.) & Knockout & 0.15 & 23.14 & 0.85 & 1.07 & 25.20 & $27.2\times$ & \textbf{+8.6\%} \\
                   & \textbf{$\Delta$} & 0\% & \textbf{+7.7\%} & \textbf{+18.1\%} & \textbf{+25.9\%} & -- & -8.7\% & -- \\
\midrule
\textbf{LogicalObj} & Normal   & 0.07 & 14.09 & 0.84 & 0.45 & 15.45 & $16.8\times$ & -- \\
(High vision dep.)  & Knockout & 0.06 & 14.63 & 0.90 & 0.49 & 16.09 & $16.3\times$ & \textbf{+4.1\%} \\
                    & \textbf{$\Delta$} & -14\% & \textbf{+3.8\%} & \textbf{+7.7\%} & \textbf{+9.8\%} & -- & -3.0\% & -- \\
\midrule
\textbf{QueryAttr} & Normal   & 0.04 & 27.98 & 0.00 & 1.00 & 29.02 & $\infty$ & -- \\
(Pure language)    & Knockout & 0.04 & 30.63 & 0.00 & 1.25 & 31.93 & $\infty$ & \textbf{+10.0\%} \\
                   & \textbf{$\Delta$} & 0\% & \textbf{+9.5\%} & 0\% & \textbf{+25.2\%} & -- & -- & -- \\
\midrule
\textbf{CompareAttr} & Normal   & 0.26 & 23.23 & 0.08 & 0.83 & 24.40 & $290\times$  & -- \\
(Extract--compare)   & Knockout & 0.26 & 23.14 & 0.01 & 0.82 & 24.24 & $2314\times$ & \textbf{-0.7\%} \\
                     & \textbf{$\Delta$} & 0\% & -0.4\% & \textbf{-86.3\%} & -1.2\% & -- & $+8\times$ & -- \\
\midrule
\textbf{ChooseAttr} & Normal   & 1.27 & 13.73 & 2.33 & 1.85 & 19.18 & $5.89\times$ & -- \\
(High redundancy)   & Knockout & 1.27 & 13.27 & 2.47 & 1.76 & 18.77 & $5.37\times$ & \textbf{-2.1\%} \\
                    & \textbf{$\Delta$} & 0\% & -3.3\% & \textbf{+6.0\%} & -5.0\% & -- & -8.8\% & -- \\
\midrule
\textbf{ChooseCat} & Normal   & 7.69 & 14.17 & 5.20 & 3.50 & 30.56 & $2.73\times$ & -- \\
(High redundancy)  & Knockout & 7.70 & 13.79 & 5.43 & 3.37 & 30.29 & $2.54\times$ & \textbf{-0.9\%} \\
                   & \textbf{$\Delta$} & +0.1\% & -2.7\% & \textbf{+4.5\%} & -3.9\% & -- & -7.0\% & -- \\
\bottomrule
\end{tabular}}
\end{table*}

\paragraph{Type~A---strong response in vision-dependent tasks.}
\textsc{ChooseRel} and \textsc{LogicalObj} validate all three predictions:
$U_V$ and $S$ increase, and the total information budget grows
(\textsc{ChooseRel}: $+8.6\%$; \textsc{LogicalObj}: $+4.1\%$).  These tasks
require sustained visual grounding, and the Image$\rightarrow$Question pathway
is a causal bottleneck for their transduction.  $U_L$ can still increase under
knockout, but only at a higher total-information cost---the hallmark of a less
efficient compensatory mechanism.

\paragraph{Type~B---compensation in language-dominated tasks.}
\textsc{QueryAttr} shows the largest $\Delta\,\text{Total}$ ($+10.0\%$) while
$U_V$ remains effectively zero: the model compensates by expanding
language-unique capacity ($\Delta U_L = +9.5\%$) and synergy
($\Delta S = +25.2\%$), confirming that even nominally language-internal
reasoning is affected when the upstream transduction channel is severed.
\textsc{CompareAttr} is atypical: $U_V$ drops sharply ($-86.3\%$) with
negligible change in total budget, consistent with an ``extract--compare''
computation where visual extraction completes early and later layers prefer
fully transduced representations.

\paragraph{Type~C---robustness in high-redundancy tasks.}
\textsc{ChooseAttr} and \textsc{ChooseCat} exhibit minimal impact
($\Delta\,\text{Total} = -2.1\%$ and $-0.9\%$).  Although $U_V$ increases
slightly, both $U_L$ and $S$ decrease, suggesting that high redundancy provides
alternative informational routes---and may even reduce cross-modal interaction
overhead when the direct transduction channel is removed.  Robustness here is
governed by \emph{multi-path redundancy} rather than pathway strength alone.

\paragraph{Task-dependence score.}
To summarise task sensitivity in a single metric, we define
$\text{Dep} := (\Delta U_V + \Delta S + \Delta\,\text{Total})\,/\,3$
(percentage units).  This score ranks \textsc{ChooseRel} as most dependent,
followed by \textsc{QueryAttr} and \textsc{LogicalObj}, while
\textsc{ChooseAttr} and \textsc{ChooseCat} are near zero or negative---consistent
with redundancy-driven robustness.

\subsubsection{Layer-wise knockout dynamics}
\label{subsubsec:knockout_layerwise}

\begin{figure*}[t]
\centering
\includegraphics[width=\linewidth]{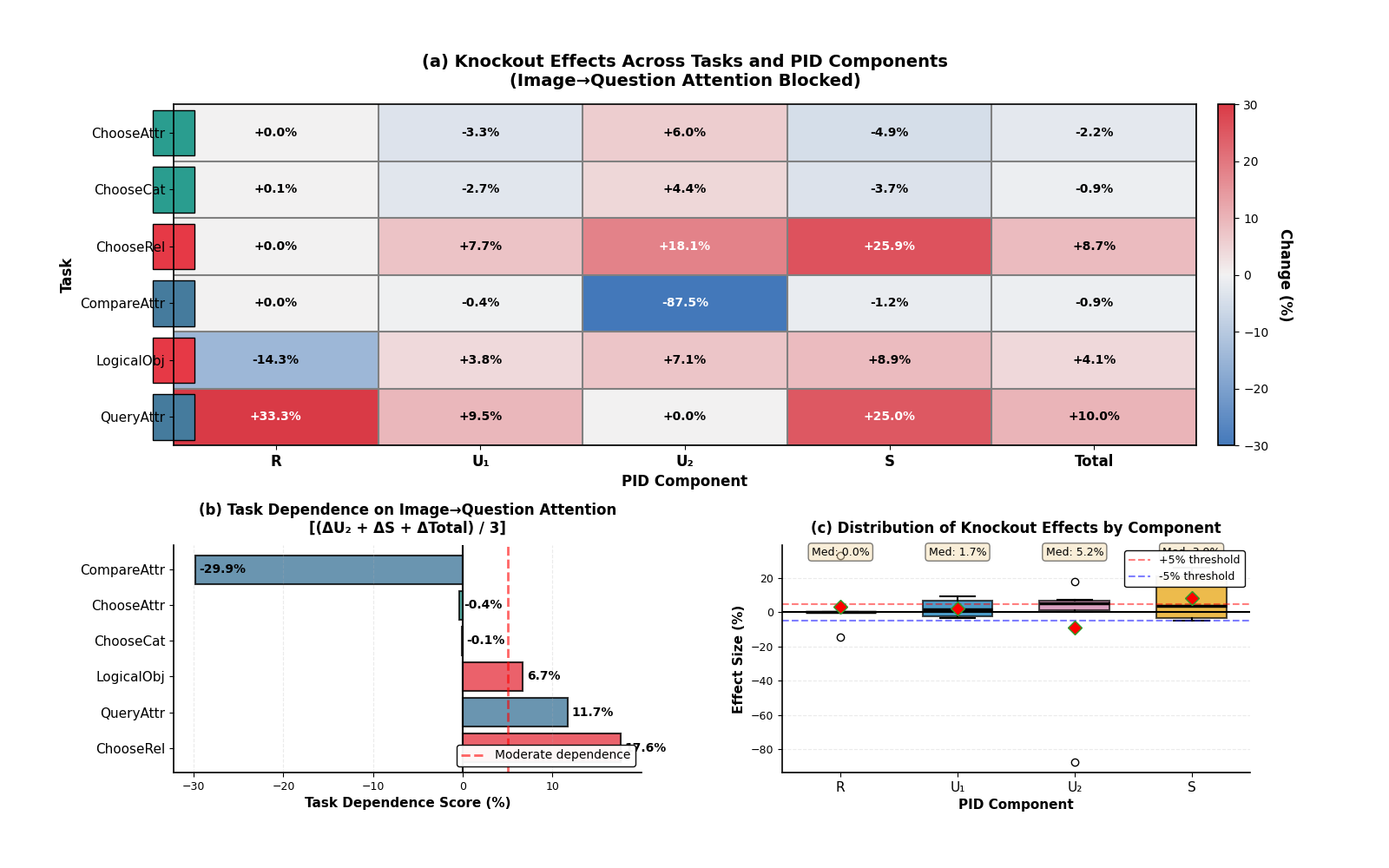}
\caption{%
\textbf{Layer-wise knockout effects for \textsc{ChooseRel}.}
Solid: normal; dashed: knockout.  Blocking Image$\rightarrow$Question attention
traps visual evidence (elevated $U_V$), delays $U_L$ growth, and forces
late-layer compensation at higher total-information cost.}
\label{fig:knockout_trajectories}
\end{figure*}

Figure~\ref{fig:knockout_trajectories} compares the full 32-layer PID
trajectories under normal and knockout conditions for \textsc{ChooseRel}, the
task with the strongest causal response.  The knockout induces a characteristic
cascade that mirrors the three-stage observational pattern from
Experiments~I--II, now with causal force:

\emph{Stage~I\,(Layers~0--10).}\;  Early layers fail to establish a strong
visually conditioned language state: $U_V$ remains elevated relative to the
baseline, and the initial $U_L$ rise is attenuated.

\emph{Stage~II\,(Layers~10--20).}\;  The gap between normal and knockout
trajectories widens as mid-layer consolidation proceeds without the full
benefit of early transduction.

\emph{Stage~III\,(Layers~20--32).}\;  Late layers compensate by sharply
increasing both $U_L$ and $S$, but at the cost of a higher total information
budget---confirming that the compensatory regime, while partially effective, is
less efficient than the intact transduction pathway.

This depth-resolved profile transforms the observational three-stage structure
into a causal account: blocking the primary transduction route at its origin
propagates through the entire network, with each stage exhibiting predictable
deviations from the unperturbed trajectory.

\medskip

In summary, attention knockout provides direct causal evidence for modal
transduction.  Blocking Image$\rightarrow$Question attention systematically
increases $U_V$ and $S$ in vision-dependent tasks and raises the total
information budget required for decision formation.  Task-level heterogeneity
reveals that semantic redundancy governs knockout robustness, and the PID lens
upgrades prior knockout findings into a quantitative account of \emph{which
information components} are carried by which pathways.

\section{Conclusion}
\label{sec:conclusion}

We introduced a layer-wise information-theoretic framework that combines
Partial Information Decomposition with scalable normalizing-flow estimation to
characterize how visual and linguistic evidence is organised across the depth
of multimodal Transformers.  Across six reasoning tasks and two LLaVA model
variants, the framework reveals a consistent \emph{modal transduction}
mechanism: visual-unique information is injected early and progressively
absorbed into language representations, language-unique information dominates
late-layer decision formation (approximately 82\% of total predictive
information at the final layer), and synergy remains bounded throughout
($<$2\%).  Causal validation via Image$\rightarrow$Question attention knockout
confirms that blocking this pathway traps visual evidence, forces compensatory
synergy, and inflates the total information budget---with effect sizes governed
by task-level semantic redundancy.

These results reframe multimodal reasoning in current MLLMs as
\emph{language-space decision making under visually conditioned
representations}, rather than emergent cross-modal fusion.  This
characterisation carries concrete design implications: if visual evidence is
compressed in early layers, improving the encoder alone will have limited
downstream impact; preserving modality-unique information likely requires
architectural or objective-level interventions---such as auxiliary losses that
penalise premature transduction, or cross-attention mechanisms that maintain a
separate visual stream---targeted at the bottleneck layers identified by PID
trajectories.

\paragraph{Limitations and future work.}
Our analysis is subject to several limitations that suggest directions for
future research.
First, the PID estimation pipeline introduces cumulative approximations
(mean pooling, PCA, flow-based Gaussianization, $I_{\min}$ redundancy);
while ordinal patterns are robust, absolute bit values should be interpreted
with caution, and future work should quantify estimation error more precisely.
Second, our experiments are restricted to the LLaVA family, which shares a
common encode-project-decode architecture; models with structurally different
fusion mechanisms (e.g., cross-attention--based architectures such as Flamingo)
may exhibit qualitatively different PID trajectories, and testing this
hypothesis is an important next step.
Third, the attention knockout is applied uniformly across all 32 layers;
layer-specific interventions would provide finer-grained causal evidence for
the three-stage transduction structure.
Finally, GQA tasks are relatively simple; more demanding benchmarks requiring
multi-step visual reasoning may reveal stronger synergistic contributions than
observed here.
We view layer-wise PID trajectories as a general-purpose diagnostic that can be
extended to these settings, and we will release all code and data to support
such extensions.

\section{Impact Statement}

This work aims to advance the understanding of multimodal reasoning in large
vision--language models through interpretability and causal analysis. The
contributions are methodological and analytical, with no direct application,
deployment, or use of sensitive data, human subjects, or automated
decision-making systems. The potential societal impacts are consistent with
those commonly associated with research on model analysis and interpretability,
and we do not identify specific ethical concerns requiring further discussion.
Generative AI tools were used solely for language editing and stylistic
refinement; all scientific content and conclusions are the authors’ own.

\bibliography{references}
\bibliographystyle{icml2026}

\newpage
\appendix
\onecolumn

\section*{Symbol Table}

\begin{table}[t]
\centering
\caption{Symbol table for key notation.}
\small
\setlength{\tabcolsep}{6pt}
\begin{tabular}{c p{0.28\linewidth} p{0.48\linewidth}}
\toprule
Symbol & Meaning & Description \\
\midrule
$V$ & Vision modality & Visual input stream (image tokens) \\
$L$ & Language modality & Linguistic input stream (question tokens) \\
$Y$ & Target variable & Prediction target (logit or probability of correct answer) \\
\midrule
$\ell$ & Layer index & Transformer layer, $\ell \in \{0,\dots,L\}$ \\
$L$ & Model depth & Total number of Transformer layers (32 in LLaVA) \\
\midrule
$\mathbf{x}_V^{(\ell)}$ & Vision representation & Aggregated visual representation at layer $\ell$ \\
$\mathbf{x}_L^{(\ell)}$ & Language representation & Aggregated language representation at layer $\ell$ \\
\midrule
$I(\cdot;\cdot)$ & Mutual information & Shannon mutual information \\
$I_{\mathrm{tot}}^{(\ell)}$ & Total predictive information & $I(\mathbf{x}_V^{(\ell)},\mathbf{x}_L^{(\ell)};Y)$ \\
\midrule
$R^{(\ell)}$ & Redundant information & Predictive information shared by vision and language \\
$U_V^{(\ell)}$ & Vision-unique information & Predictive information available only from vision \\
$U_L^{(\ell)}$ & Language-unique information & Predictive information available only from language \\
$S^{(\ell)}$ & Synergistic information & Predictive information available only jointly \\
\midrule
$\mathcal{I}^{(\ell)}$ & Information state & $(R^{(\ell)}, U_V^{(\ell)}, U_L^{(\ell)}, S^{(\ell)})$ \\
$\mathcal{T}$ & PID trajectory & Layer-wise sequence $\{\mathcal{I}^{(\ell)}\}_{\ell=0}^L$ \\
\midrule
$\Delta C^{(\ell)}$ & Knockout-induced change & Relative change of component $C$ under attention knockout \\
\bottomrule
\end{tabular}
\end{table}

\section{Notation Summary}
\label{app:notation}

This appendix summarizes the notation used throughout the paper to ensure
clarity and consistency, especially for information-theoretic quantities and
layer-wise analysis in multimodal large language models (MLLMs).

\paragraph{Modalities and representations.}
We denote the vision modality by $V$ and the language modality by $L$.
At Transformer layer $\ell$, the aggregated modality-level representations are
$\mathbf{x}_V^{(\ell)}$ for vision and $\mathbf{x}_L^{(\ell)}$ for language.
Aggregation is performed over modality-specific token sets as described in
Section~\ref{sec:method}.

\paragraph{Target variable.}
The target variable $Y$ denotes the model’s prediction target and is implemented
as the logit (or logit-transformed probability) of the correct answer token.
All information quantities are computed with respect to $Y$.

\paragraph{Partial Information Decomposition (PID).}
At each layer $\ell$, the joint predictive information is decomposed as
\begin{equation}
I(\mathbf{x}_V^{(\ell)}, \mathbf{x}_L^{(\ell)}; Y)
= R^{(\ell)} + U_V^{(\ell)} + U_L^{(\ell)} + S^{(\ell)} ,
\end{equation}
where $R^{(\ell)}$ is redundant information shared by both modalities,
$U_V^{(\ell)}$ and $U_L^{(\ell)}$ are vision-unique and language-unique
information, respectively, and $S^{(\ell)}$ is synergistic information that
arises only from the joint pair.

\paragraph{Information state and trajectory.}
We define the layer-wise information state as
$\mathcal{I}^{(\ell)} = (R^{(\ell)}, U_V^{(\ell)}, U_L^{(\ell)}, S^{(\ell)})$,
and the PID trajectory as the sequence
$\mathcal{T} = \{\mathcal{I}^{(\ell)}\}_{\ell=0}^{L}$.

\paragraph{Total information.}
The total predictive information at layer $\ell$ is denoted by
\[
I_{\mathrm{tot}}^{(\ell)} := R^{(\ell)} + U_V^{(\ell)} + U_L^{(\ell)} + S^{(\ell)} .
\]

\paragraph{Attention knockout notation.}
For attention knockout experiments, we denote the relative change of a PID
component $C \in \{R, U_V, U_L, S, I_{\mathrm{tot}}\}$ at layer $\ell$ by
\[
\Delta C^{(\ell)}(\%) =
\frac{C_{\mathrm{KO}}^{(\ell)} - C_{\mathrm{Base}}^{(\ell)}}
{C_{\mathrm{Base}}^{(\ell)}} \times 100\% ,
\]
where $\mathrm{Base}$ and $\mathrm{KO}$ refer to the normal and knockout
conditions, respectively.

\paragraph{Notation consistency.}
Throughout the paper, we use $(V,L)$ consistently to denote vision and language
modalities. Any alternative labels (e.g., question/image tokens) refer to the
same underlying distinction and are unified under this notation.

\section{Experimental Settings and Implementation Details}
\label{sec:app_settings}

\subsection{Models and Comparative Design}
\label{app:model}

We study the \textsc{LLaVA} family, which instantiates the dominant multimodal design of a \emph{visual encoder} followed by a \emph{projection} into an \emph{LLM backbone}. Importantly, \textsc{LLaVA} maintains an explicit separation between visual and textual token streams, which makes the modality-level random variables required by our PID analysis well-defined.

Our main comparison contrasts \textsc{LLaVA-1.5-7B} and \textsc{LLaVA-1.6-7B} (Vicuna-based). Both models use a 32-layer Transformer language backbone, enabling layer-wise alignment at the level of representational hierarchy, while differing meaningfully in visual processing. \textsc{LLaVA-1.5-7B} uses CLIP ViT-L/14 with fixed resolution ($336{\times}336$), yielding a fixed number of visual tokens ($576$). These tokens are mapped by a two-layer MLP into the Vicuna embedding space (dimension $4096$; Vicuna-v1.5-7B). In contrast, \textsc{LLaVA-1.6-7B} supports dynamic high-resolution and multi-scale processing (up to $672{\times}672$), producing a variable number of visual tokens (typically $576$--$2304$), and is trained on an expanded dataset with an updated Vicuna version (v1.6). This controlled perturbation allows us to test whether the observed information-flow patterns reflect architecture-specific details (e.g., resolution and token count) or an \emph{architecture-robust} mechanism driven by LLM priors and task semantics.

\subsection{Tasks, Semantic Coverage, and Sample Scale}
\label{app:tasks}

We conduct our experiments on the GQA dataset~\cite{hudson2019gqa}, a large-scale benchmark for real-world visual reasoning and compositional question answering built on images from Visual Genome~\cite{krishna2017visual}. GQA provides scene-graph annotations with objects, attributes, and relations, and uses a functional program engine to generate diverse yet controlled questions with reduced language bias and balanced answer distributions. Its questions are organized along structural and semantic dimensions, covering existence, attribute, category, relation, and logical queries, which makes it particularly suitable for probing how multimodal large language models integrate visual and linguistic information during visual question answering.

\paragraph{Our GQA-based VQA subset.}
Starting from the validation split of GQA, we construct a focused VQA subset tailored to our analysis. We keep question groups where state-of-the-art MLLMs such as LLaVA variants achieve reasonably high accuracy, and discard overly easy yes/no verification questions. From six remaining groups that span attribute and category selection, spatial reasoning, and object-level comparison or logical inference, we sample image–question pairs that are correctly answered by the models under study. This results in a compact benchmark with reliable model predictions and object-level bounding boxes, which we exploit to separate object-related patches from the rest of the image when studying cross-modal information flow.

We construct six representative visual reasoning subtasks from the GQA validation set:
\textsc{ChooseAttr}, \textsc{ChooseCat}, \textsc{ChooseRel}, \textsc{CompareAttr},
\textsc{LogicalObj}, and \textsc{QueryAttr}. Collectively, the suite covers attribute recognition, category identification, spatial relations, attribute comparison, logical composition, and spatial localization.

The task suite is designed to (i) provide semantic diversity without duplicating the same reasoning template, and (ii) induce meaningful variation in information structure. In particular, tasks differ in visual dependence (e.g., spatial relations and logical composition typically require stronger grounding) and cross-modal redundancy (e.g., category concepts often have strong shared representations in vision and language). For each task and model, we retain at least $900$ valid samples, which is sufficient for stable layer-wise estimation and reduces finite-sample bias in mutual-information proxies.

\section{Layer-wise PID Flow Estimation and Gaussian Decomposition}
\label{sec:app_pidflow}

\subsection{Why Layer-wise Independent Estimation}
\label{app:layerwise}

Our object of interest is the predictive information structure at each layer. Because the representation distribution changes substantially with depth, we train PID-flow estimators independently for each layer. Sharing a single estimator across layers would introduce coupling artifacts and bias trajectory comparisons. For each task, model, and layer, we train two bivariate flows: one for $(X_L^\ell,Y)$ and one for $(X_V^\ell,Y)$.

\subsection{PID Flow Objective and Operational Semantics}
\label{app:pidflow}

For a fixed layer $\ell$, we learn invertible mappings
\begin{equation}
(\hat Z_Q,\hat Z_Y)=f_Q^\ell(X_L^\ell,Y),\qquad
(\hat Z_I,\hat Z_Y')=f_I^\ell(X_V^\ell,Y),
\end{equation}
such that each transformed pair is approximately standard Gaussian. This enables closed-form Gaussian mutual information and a numerically stable PID proxy satisfying non-negativity and additivity. We emphasize that these PID terms are \emph{operational proxies} used to characterize trajectories and intervention responses, rather than claims of exact ground-truth PID in the underlying high-dimensional system.

\subsection{Flow Architecture and Optimization}
\label{app:flow}

We use RealNVP with 8 coupling blocks per bivariate flow, hidden dimension 256, ReLU activations, and batch normalization after each block. We optimize with Adam (learning rate $10^{-4}$, batch size 128, 10{,}000 steps), weight decay $10^{-5}$, and gradient clipping at $1.0$. All runs use fixed random seeds and CUDA deterministic mode.

\subsection{Gaussian MI and PID Construction}
\label{app:gaussian}

For a Gaussian pair $(A,B)$ with correlation $\rho_{AB}$, mutual information is
\begin{equation}
I(A;B)=-\tfrac{1}{2}\log(1-\rho_{AB}^2).
\end{equation}
Using estimated correlations, we compute $I(Z_Q;Z_Y)$, $I(Z_I;Z_Y)$, and $I(Z_Q,Z_I;Z_Y)$. We define redundancy by the minimum-information rule:
\begin{equation}
R^\ell=\min\{I(Z_Q;Z_Y),\,I(Z_I;Z_Y)\},
\end{equation}
and the remaining components by additivity:
\begin{equation}
U_L^\ell=I(Z_Q;Z_Y)-R^\ell,\quad
U_V^\ell=I(Z_I;Z_Y)-R^\ell,\quad
S^\ell=I(Z_Q,Z_I;Z_Y)-I(Z_Q;Z_Y)-I(Z_I;Z_Y)+R^\ell.
\end{equation}
By construction, $R^\ell+U_L^\ell+U_V^\ell+S^\ell=I(Z_Q,Z_I;Z_Y)$ up to numerical tolerance.

\subsection{Numerical Stability and Quality Checks}
\label{app:stability}

For each task--model--layer configuration, we verify (i) non-negativity of all PID components, (ii) additivity error $<1\%$, and (iii) seed stability via repeated training. We focus on robust properties---relative dominance, turning points, and trajectory trends---and report variance ranges to ensure that the observed patterns reflect mechanisms rather than estimator noise.

\section{Cross-Model Consistency Metrics}
\label{app:consistency}

To quantify architecture robustness, we align PID trajectories between \textsc{LLaVA-1.5} and \textsc{LLaVA-1.6} for each task and component. We compute Pearson correlation across layers and normalized mean absolute error (nMAE). We additionally identify key layers (e.g., dominance switches or curvature peaks) and report their absolute offsets, localizing any discrepancies to specific components and depth ranges.

\section{Attention Knockout: Implementation and Paired Evaluation}
\label{app:knockout}

\subsection{Knockout Design}
\label{app:knockout_design}

We implement a unidirectional knockout that blocks \emph{Image$\rightarrow$Question} attention by setting the corresponding attention logits to $-\infty$ before softmax, while preserving \emph{Question$\rightarrow$Image} and all self-attention pathways. The intervention is applied simultaneously to all 32 layers, targeting the hypothesized modal-transduction route.

\subsection{Mechanistic Intervention and Paired Computation}
\label{app:knockout_paired}

The knockout modifies only the forward computation graph and leaves parameters unchanged, constituting a clean mechanistic intervention. We evaluate knockout and baseline on identical samples (as in Section~\ref{app:tasks}), enabling paired comparisons.

\subsection{PID Under Knockout and Task-level Dependence}
\label{app:knockout_pid}

Under knockout, we recompute PID trajectories even when predictions become incorrect, enabling analysis of how information structure reorganizes under disruption. We report relative changes
\begin{equation}
\Delta C^\ell(\%)=\frac{C^\ell_{\mathrm{KO}}-C^\ell_{\mathrm{Base}}}{C^\ell_{\mathrm{Base}}}\times100\%,
\end{equation}
and define a task-level dependence score as the average of $\Delta U_V$, $\Delta S$, and $\Delta I_{\text{Total}}$.

\section{Evaluation, Statistics, and Reproducibility}
\label{app:stats}

We report absolute PID values (bits) as well as normalized proportions. Key summary metrics include the final-layer ratio $U_L^L/U_V^L$, the synergy share $S^L/I_{\text{Total}}^L$, and cross-layer turning points. Statistical stability is assessed using bootstrap confidence intervals (1{,}000 resamples), ANOVA across tasks, and paired $t$-tests for knockout effects, reporting Cohen's $d$ as effect size.

All experiments use fixed random seeds (42), deterministic computation, fixed data splits, and disjoint sets for flow training and PID evaluation. We will release code, preprocessing scripts, and visualization pipelines upon publication to support full reproducibility.

\section{Supplementary Material for Layer-wise PID Analysis}
\label{app:pid_supplement}

This appendix provides supplementary material for the method introduced in
Section~\ref{sec:method}. It consolidates robustness checks, proof sketches,
and implementation details that support the theoretical and empirical validity
of the proposed layer-wise PID analysis framework.

\subsection{Alternative Pooling Rules and Robustness}
\label{app:pooling}

In the main text, modality-level representations are summarized using mean
pooling over visual and language token regions
(Section~\ref{sec:layerwise_state}). Here we assess the robustness of PID
trajectories to alternative pooling strategies.

We consider (i) attention-weighted pooling over language tokens using the
self-attention distribution, and (ii) max pooling over token dimensions.
Across all tasks and models, the qualitative behavior of PID trajectories is
preserved, including the relative dominance of $U_V^{(\ell)}$, $U_L^{(\ell)}$,
and $S^{(\ell)}$, as well as the location of turning points.
While absolute magnitudes vary mildly, all mechanism classifications defined
in Section~\ref{sec:three_mechanisms} remain unchanged.

\subsection{Separability of Mechanism Definitions}
\label{app:separability}

This section provides a proof sketch for
Proposition~\ref{prop:mechanism_separability}.

Let $\mathcal{I}^{(L)} = (R^{(L)}, U_V^{(L)}, U_L^{(L)}, S^{(L)})$ denote the
final-layer information allocation, and assume
$I_{\mathrm{tot}}^{(L)} > 0$.
Each mechanism definition in Section~\ref{sec:three_mechanisms} imposes a set of
strict inequality constraints on $\mathcal{I}^{(L)}$.
Persistent synergy requires proportional dominance of $S^{(L)}$;
modal transduction requires dominance of $U_L^{(L)}$ together with an
intermediate-layer peak in $U_V^{(\ell)}$;
and redundancy-dominant convergence requires $R^{(L)}$ to exceed all other
components by a fixed margin.
Under non-degeneracy (at least one component strictly dominates), these regions
do not overlap in the simplex defined by
$R^{(L)} + U_V^{(L)} + U_L^{(L)} + S^{(L)} = I_{\mathrm{tot}}^{(L)}$.

\subsection{Sensitivity to PCA Dimensionality}
\label{app:pca_sensitivity}

We evaluate the sensitivity of PID trajectories to the retained PCA dimension
$d'$.
Across $d' \in \{16, 32, 64\}$, corresponding to approximately
$90\%$--$98\%$ variance retained, the qualitative structure of PID trajectories
is stable.
In particular, the ordering of modality dominance at the final layer and the
depth of modality-transduction turning points are preserved.
This suggests that PCA primarily removes redundant noise rather than
decision-critical information.

\subsection{Gaussian Plug-in PID: Closed-form Details}
\label{app:gaussian_pid}

After the flow transformation described in
Section~\ref{sec:pid_flow}, we model
$(\mathbf{z}_V^{(\ell)}, \mathbf{z}_L^{(\ell)}, Y)$ as jointly Gaussian.
For Gaussian variables, mutual information admits the closed form
\begin{equation}
I(A;B) = -\tfrac{1}{2}\log\!\left(1 - \rho_{AB}^2\right),
\end{equation}
where $\rho_{AB}$ denotes the Pearson correlation coefficient.
All marginal and joint mutual information terms are computed from empirical
covariance estimates.
PID components are then obtained via the defining relations in
Section~\ref{sec:prelim_pid}, adopting the $I_{\min}$ redundancy rule.
Numerical non-negativity is enforced up to a tolerance of $10^{-6}$.

\subsection{Consistency of Gaussian Plug-in Estimation}
\label{app:consistency}

Under correct model specification, i.e., when the transformed variables are
exactly Gaussian, maximum-likelihood covariance estimation is consistent.
By the continuous mapping theorem, the induced mutual information estimates and
the resulting PID components are consistent as the sample size tends to
infinity.
When Gaussianity is violated, bias may arise; in practice, we mitigate this via
flow-capacity sweeps, normality diagnostics, and cross-model consistency checks.

\subsection{Threshold Sensitivity}
\label{app:threshold_sensitivity}

All thresholds introduced in Section~\ref{sec:three_mechanisms}
(e.g., $\tau_S$, $\gamma$, $\eta$, and $\rho$) are varied within reasonable
ranges.
The qualitative classification of mechanisms remains stable across these
ranges, indicating that our conclusions are not driven by finely tuned
hyperparameters.

\end{document}